\documentclass{article}

\usepackage{microtype}
\usepackage{graphicx}
\usepackage{subcaption}
\usepackage{booktabs} % for professional tables

\usepackage{hyperref}

\usepackage[preprint]{icml2026}

\usepackage{amsmath}
\usepackage{amssymb}
\usepackage{mathtools}
\usepackage{amsthm}
\usepackage{multirow}
\usepackage{diagbox}

\usepackage[capitalize,noabbrev]{cleveref}

\theoremstyle{plain}

\theoremstyle{definition}

\theoremstyle{remark}

\usepackage[textsize=tiny]{todonotes}

\icmltitlerunning{Not all tokens contribute equally to diffusion learning}

\usepackage[table]{xcolor}

\definecolor{IceBlue}{rgb}{0.88,0.95,1.0}
\definecolor{CalmBlue}{rgb}{0.80,0.88,0.98}
\definecolor{AcademicBlue}{rgb}{0.75,0.85,0.95}
\definecolor{SlateBlue}{rgb}{0.70,0.78,0.90}
\definecolor{SkyMist}{rgb}{0.85,0.93,0.98}

\definecolor{MintGreen}{rgb}{0.88,1.0,0.94}
\definecolor{SoftTeal}{rgb}{0.78,0.92,0.88}
\definecolor{PaleGreen}{rgb}{0.86,0.95,0.86}
\definecolor{AquaGreen}{rgb}{0.80,0.90,0.85}
\definecolor{NatureGreen}{rgb}{0.74,0.86,0.78}

\definecolor{CreamYellow}{rgb}{1.0,0.98,0.85}
\definecolor{SoftGold}{rgb}{0.98,0.92,0.70}
\definecolor{LightAmber}{rgb}{1.0,0.94,0.80}
\definecolor{WarmSun}{rgb}{0.96,0.90,0.75}
\definecolor{Apricot}{rgb}{1.0,0.92,0.85}

\definecolor{Lavender}{rgb}{0.93,0.90,0.98}
\definecolor{LightLilac}{rgb}{0.95,0.88,0.98}
\definecolor{SoftPink}{rgb}{1.0,0.88,0.92}
\definecolor{RoseMist}{rgb}{0.98,0.90,0.94}
\definecolor{BlushPink}{rgb}{0.97,0.85,0.88}

\definecolor{NatureBlue}{rgb}{0.36,0.54,0.66}
\definecolor{NatureOrange}{rgb}{0.83,0.56,0.33}
\definecolor{NaturePurple}{rgb}{0.65,0.48,0.62}
\definecolor{NatureGray}{rgb}{0.70,0.70,0.70}

\definecolor{LightGray}{rgb}{0.93,0.93,0.93}
\definecolor{SoftBlue}{rgb}{0.88,0.94,1.0}
\definecolor{SoftGreen}{rgb}{0.88,1.0,0.88}
\definecolor{SoftYellow}{rgb}{1.0,1.0,0.88}
\definecolor{SoftPink}{rgb}{1.0,0.88,0.88}
\definecolor{PaleOrange}{rgb}{1.0, 0.95, 0.85} % 浅橙色
\definecolor{SkyBlue}{rgb}{0.8, 0.9, 1.0}      % 天蓝色

\usepackage{pifont}

\newcommand{\gain}[1]{{\color{green!50!black}\scriptsize(\ding{58} \textbf{#1})}}

\begin{document}

\twocolumn[
  \icmltitle{Not all tokens contribute equally to diffusion learning}



  \icmlsetsymbol{corresponding}{*}
  \icmlsetsymbol{equal}{$^\dagger$}

  \begin{icmlauthorlist}
    \icmlauthor{Guoqing Zhang}{equal,1,2,3}
    \icmlauthor{Lu Shi}{1,2,3}
    \icmlauthor{Wanru Xu}{1,2,3}
    \icmlauthor{Linna Zhang}{4}
    \icmlauthor{Sen Wang}{5}
    \icmlauthor{Fangfang Wang}{corresponding,6}
    \icmlauthor{Yigang Cen}{corresponding,1,2,3}
  \end{icmlauthorlist}

  \icmlaffiliation{1}{State Key Laboratory of Advanced Rail Autonomous Operation, Bejing Jiaotong University, Beijing, China}
  \icmlaffiliation{2}{School of Computer Science and Technology, Bejing Jiaotong University, Beijing, China}
  \icmlaffiliation{3}{Visual Intellgence +X International Cooperation Joint Laboratory of MOE, Bejing Jiaotong University, Beijing, China}
  \icmlaffiliation{4}{School of Mechanical Engineering, Guizhou University, Guizhou, China}
  \icmlaffiliation{5}{Seed, Bytedance, Beijing, China}
  \icmlaffiliation{6}{School of Information Science and Technology, Hangzhou Normal University, Zhejiang, China}

  \icmlcorrespondingauthor{Guoqing Zhang}{guoqing.zhang@bjtu.edu.cn}
  
  \icmlkeywords{Machine Learning, ICML}

  \vskip 0.3in
]



\printAffiliationsAndNotice{}  % no special notice (required even if empty)

\begin{abstract}
With the rapid development of conditional diffusion models, significant progress has been made in text-to-video generation. However, we observe that these models often neglect semantically important tokens during inference, leading to biased or incomplete generations under classifier-free guidance. We attribute this issue to two key factors: distributional bias caused by the long-tailed token frequency in training data, and spatial misalignment in cross-attention where semantically important tokens are overshadowed by less informative ones.
To address these issues, we propose \textbf{D}istribution-\textbf{A}ware \textbf{R}ectification and Spatial \textbf{E}nsemble (DARE), a unified framework that improves semantic guidance in diffusion models from the perspectives of distributional debiasing and spatial consistency. First, we introduce Distribution-Rectified Classifier-Free Guidance (DR-CFG), which regularizes the training process by dynamically suppressing dominant tokens with low semantic density, encouraging the model to better capture underrepresented semantic cues and learn a more balanced conditional distribution. This design mitigates the risk of the model distribution overfitting to tokens with low semantic density. Second, we propose Spatial Representation Alignment (SRA), which adaptively reweights cross-attention maps according to token importance and enforces representation consistency, enabling semantically important tokens to exert stronger spatial guidance during generation. This mechanism effectively prevents low semantic-density tokens from dominating the attention allocation, thereby avoiding the dilution of the spatial and distributional guidance provided by high semantic-density tokens.
Extensive experiments on multiple benchmark datasets demonstrate that DARE consistently improves generation fidelity and semantic alignment, achieving significant gains over existing approaches.
\end{abstract}

\section{Introduction}
\label{sec:intro}
With the continuous advancement of methods for text-to-image (T2I) generation \cite{sd35,flux.1,sana}, substantial progress has been made in text-to-video (T2V) generation, which has rapidly emerged as another core task in generative modeling. The primary objective of T2V is to enable models to synthesize temporally coherent and semantically consistent videos conditioned on natural language prompts. Early works \cite{make-a-video,pixeldance} primarily focused on optimizing U-Net–based \cite{unet} diffusion model architectures to enhance temporal consistency and visual coherence in generated videos. More recent approaches, represented by Wan \cite{wan}, CogVideoX \cite{Cogvideox}, and Hunyuan-Video \cite{Hunyuanvideo}, adopt Transformer-based \cite{transformer} diffusion models as the backbone architecture and improve video generation quality from perspectives such as spatiotemporal compression ratios, sparse representations, and semantic–visual alignment. Although these methods achieve significant improvements in video quality by optimizing visual encoders and diffusion model architectures, they largely overlook the impact of semantic representations on the T2V task and on diffusion model design.

\begin{figure}[t]
\centering
  
\begin{subfigure}{\linewidth}
    \centering
    \includegraphics[width=0.99\linewidth]{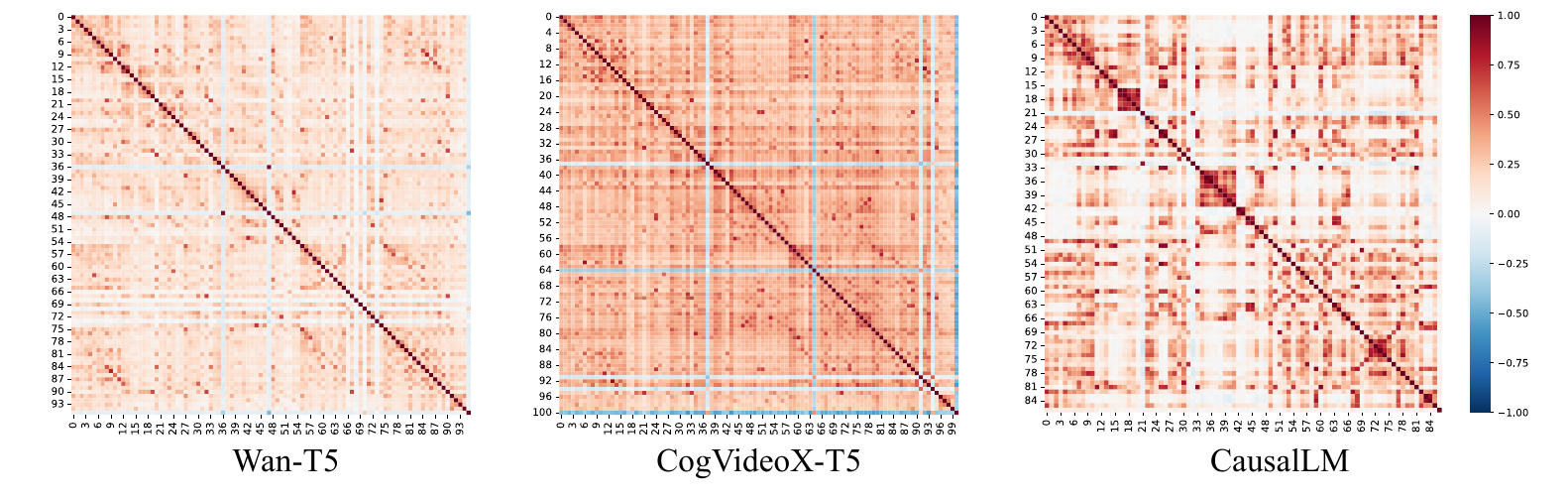}
    \caption{Visualization of Similarity between Conditional Semantic Representations in Diffusion Model Inputs.} 
    \label{fig:img_embedding_sim}
\end{subfigure}
\begin{subfigure}{\linewidth}
    \centering
    \includegraphics[width=0.99\linewidth]{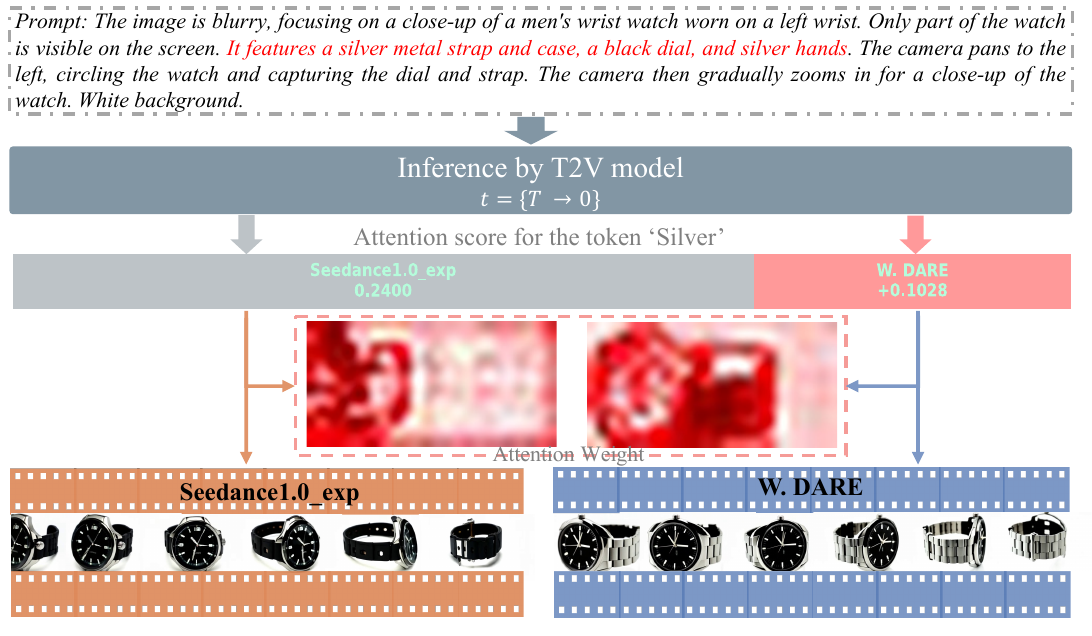}
    \caption{Visualization of Attention Scores and Distributions for Underfitted Tail Tokens During Inference}
    \label{fig:long_tail_token}
\end{subfigure}
  
  \caption{Visualization and Analysis of Conditional Semantic Information Distribution. (a) shows the similarity between conditional semantic representations. The off-diagonal regions indicate that the semantic similarity between different tokens is low, reflecting a discrete distribution. (b) presents the attention scores and corresponding distribution maps of selected underfitted tail tokens generated by Seedance \cite{seedance} before and after the introduction of DARE.}
  \label{fig:intro}
  \vspace{-15pt} 
\end{figure}

%  图1想说明我们一开始抛出的“生成效果不够好的原因在于模型attend错token”这件事是一个事实，但（b）只是从token分布和结果上对得上我们说的这个问题，好像不足以说明推理过程的岔子出在哪。直接用token activation map或者额外加上tam会不会更有说服力？

First, both traditional approaches and recent studies typically rely on text encoders such as T5 \cite{t5,Glyph-byt5}, CLIP \cite{clip}, and Qwen \cite{qwen} to extract semantic representations during semantic processing. However, we observe a consistent phenomenon across different architectures. Whether using T5, which is pre-trained with a conventional encoder–decoder architecture, or Qwen, which is pre-trained with an autoregressive decoder architecture, the encoded semantic representations of individual tokens remain highly discrete. As illustrated in Fig.~\ref{fig:img_embedding_sim}, the semantic similarity between different tokens is relatively low, and the representations tend to be mutually independent.
Second, during training, the increase in training data leads to the frequent occurrence of tokens carrying low-density semantic information, such as conjunctions and prepositions (e.g., the, to, and an), which further exacerbates the imbalance in token distribution. Considering that the semantic information carried by different tokens is relatively independent, the model tends to fit the distribution of high-frequency tokens with low semantic density, thereby weakening its ability to accurately model the distribution of tokens that contain high-density semantic information. As illustrated in Fig.~\ref{fig:long_tail_token}, by extracting the attention scores and attention maps corresponding to tail tokens during inference, we can observe that these tokens generally receive relatively low attention scores and attend to spatially scattered regions, indicating insufficient learning. Consequently, when classifier-free guidance is additionally applied during inference for video generation, this bias is further amplified, reinforcing constraints associated with low-frequency semantic tokens and leading to the loss of certain semantic information in the generated content.

To address the above issues, we propose Distribution-Aware Rectification and Spatial Ensemble (DARE), a unified framework that improves semantic guidance in conditional diffusion models from two complementary perspectives: distributional rectification and spatial representation alignment. First, from the perspective of distribution rectification, we propose Distribution-Rectified Classifier-Free Guidance (DR-CFG). Instead of modifying the guidance during inference, DR-CFG regularizes the learning process during training by randomly dropping or weakening a subset of frequently occurring and easily fitted dominant tokens, preventing the model from overemphasizing tokens with low semantic density. Specifically, we selectively identify tokens with low semantic density and compute their corresponding flow distributions. The resulting flow distribution is then contrasted with the flow distribution obtained under the guidance of the complete phrase. The differential information is further used to dynamically adjust the flow matching loss distribution in diffusion models. This strategy reduces the excessive influence of tokens carrying low-density semantic information on the training objective, while encouraging the model to better capture semantic cues that are otherwise underrepresented. As a result, the learned conditional distribution becomes more balanced and robust.
Second, from the perspective of spatial representation alignment, we introduce a Spatial Representation Alignment (SRA) method. This method dynamically regulates the attention distribution during training, allocating more attention to tokens with high semantic density while suppressing the dominance of low-density tokens, thereby preventing semantically informative tokens from being overshadowed in the attention allocation. Concretely, we dynamically reweight attention scores based on token importance, suppressing the dominance of low-semantic-density tokens in the attention distribution while enhancing the spatial contribution of high-density semantic tokens. A consistency loss is further employed to align the reweighted attention outputs with the original attention representations. This encourages the model to learn semantically aware spatial representations, enabling it to better capture the spatial guidance signals provided by high-density semantic tokens.
Finally, we evaluated the generation performance of our method on multiple benchmark datasets and performed comparative experiments against existing approaches. The results demonstrate that our method consistently and significantly improves both the generation capability and the generation quality of the base model.

% \begin{figure}[t]
% \centering
% \begin{subfigure}[t]{\linewidth}
%     \centering
%     \includegraphics[width=\linewidth]{images/intro_a_img_embedding.pdf}
%     \caption{Visualization of Similarity between Conditional Semantic Representations in Diffusion Model Inputs.}
%     \label{fig:img_embedding_sim}
% \end{subfigure}
% \begin{subfigure}[t]{\linewidth}
%     \centering
%     \includegraphics[width=\linewidth]{images/intro_b_longtail.pdf}
%     \caption{The frequency distribution of different tokens during training, along with the inference results, after encoding the conditional text with a tokenizer.}
%     \label{fig:long_tail_token}
% \end{subfigure}
% \caption{Visualization and Analysis of Conditional Semantic Information Distribution. (a) shows the similarity between conditional semantic representations. The off-diagonal regions indicate that the semantic similarity between different tokens is low, reflecting a discrete distribution. (b) illustrates the frequency of different token occurrences during the training phase. Due to the relatively discrete and independent nature of the semantic representations, the model tends to overlook tokens with high semantic information but low frequency during the inference phase, leading to relatively poor generation results.}
% \label{fig:intro}
% \end{figure}

Our key contributions are as follows:
\begin{itemize}
\item This paper introduces the Distribution-Rectification Classifier-Free Guidance (DR-CFG) method, which, for the first time, corrects the model's preference distribution from the perspective of tokens. This effectively enhances the model's ability to fit high-semantic information tokens.

\item We propose the Spatial Representation Alignment (SRA) method, which uses a self-supervised learning framework to dynamically adjust the spatial distribution of features based on the semantic density of tokens. This improves the consistency between spatial representations and high-semantic information tokens.

\item By integrating our methods with existing video generation models, we significantly improve the video generation quality of the base model and achieve optimal performance across multiple benchmark tests.
\end{itemize}

\section{Related Works}

\subsection{Text-to-Video Generation}
The Text-to-Video (T2V) task aims to generate controllable videos from textual semantics. Early works primarily adopted GAN-based architectures \cite{gan_t2v,TGans_c}, incorporating text into the generator and enforcing semantic consistency via discriminators, but often suffered from training instability. More recently, diffusion-based frameworks have emerged as a more stable and effective alternative \cite{svd}. For example, Make-A-Video \cite{make-a-video} incorporates a spatiotemporal module into a UNet-based T2I model, reducing reliance on large text-video datasets, while Animatediff \cite{animatediff} leverages LoRA-based fine-tuning to generate short animations. Methods such as MagicVideo \cite{Magicvideo}, LAVIE \cite{Lavie}, and PixelDance \cite{pixeldance} further improve temporal and semantic consistency via 3D U-Nets, cascaded latent models, and conditional noise perturbation strategies. Transformer-based video diffusion models, including CogVideoX \cite{Cogvideox}, HunyuanVideo \cite{Hunyuanvideo}, and WAN \cite{wan}, extend these approaches to long, high-resolution videos using 3D VAEs, sparse attention, and large-scale Transformer backbones, enabling multi-resolution, high-quality video generation.

% 建议在2.1和2.2小节的最后用简洁两句话说一下我们方法在literature里的定位和独特的亮点

\subsection{Classifier-Free Guidance}
% The classifier-free guidance (CFG) learning mechanism was first introduced in \cite{cfg}, which enables effective conditional guidance on specific attributes or categories without the need for an additional classifier. This mechanism improves generation quality and provides better control over the generated content. Since then, more works have incorporated CFG into both model training and inference phases to enhance generative performance.
The classifier-free guidance (CFG) mechanism was first introduced in \cite{cfg}, enabling effective conditional generation without requiring an auxiliary classifier. Since then, many works have extended CFG to improve generative performance. S-CFG \cite{scfg} combines the attention distribution of the base model with object mask information during inference to adjust the feature distribution under the CFG strategy, mitigating the issue of distribution imbalance across different regions. NPO \cite{npo} constructs positive and negative text guides to calculate the model's predicted distributions and strengthens the conditional guidance of the positive text samples through distributional differences. ICG \cite{icg} builds perturbation weights based on time-step values to dynamically adjust the CFG guidance weight. SEG \cite{seg} modifies the CFG guidance scale by controlling the energy curvature of attention, thereby improving the stability and quality of image generation. NAG \cite{nag} introduces the CFG calculation method directly into the attention mechanism. By applying weighted modulation to the attention features guided by both positive and negative samples, it enhances the feature representation under positive conditional guidance, thereby improving the model's generation quality.

Unlike existing approaches that modify guidance signals during inference, we identify the root cause of semantic degradation as the distribution imbalance among tokens during training, which leads to partial loss of semantic information in the generated content. Based on the above analysis, we propose an explicit modeling strategy during training to address two key issues: (1) the fitting bias introduced by the long-tailed token distribution, and (2) the suppression of high-semantic-density tokens in terms of attention intensity and gradient contribution. Specifically, we dynamically adjust both gradient contributions and attention scores according to the fitting status of different tokens, thereby enhancing the learning and distribution fitting of high-semantic-density tokens.
With this approach, the model is encouraged to learn more comprehensive semantic representations across all tokens. As a result, it effectively alleviates the degradation in generation quality caused by insufficient learning of high-semantic-density tokens, which otherwise leads to inadequate guidance from relevant tokens during inference.

\section{Method}
\subsection{Preliminaries}
Diffusion models typically construct a stochastic differential equation (SDE) or a Markov chain to gradually transform the data distribution toward an easy-to-sample prior distribution (e.g., a standard Gaussian).
\begin{align}
	dx=f(x,t)dt+g(t)dw
\label{eq:sde_eq}
\end{align}
Here, $dx$ represents the Brownian motion corresponding to an infinitesimal Gaussian distribution, $(f, g(t))$ denotes the functions that determine the noise injection method based on the noise addition strategy \cite{elu_noise_schedule,score_noise_schedule}, and $dw$ indicates the standard Wiener process. Their training objective is usually to estimate the noise or the score (gradient field), and samples are generated by simulating the corresponding reverse-time process.

In contrast, Flow Matching \cite{flow_matching} is formulated as a deterministic ordinary differential equation (ODE) without inherent stochasticity. As a result, it is usually more efficient at sampling and can achieve high-quality generation with significantly fewer steps. Flow Matching directly learns a vector field $v_t(x)$ that matches the true generative direction along the target probability path, without the need to simulate explicit noise processes or to compute Jacobian-induced terms of the log probability density. Its forward process uses simple linear interpolation to compute $x_t = (1-t)x + t\epsilon $, where $x$ is the pure data, $t$ is the time step, and $\epsilon$ is the noise. The derivative of $x$ with respect to $t$ at time $t$ is given by $\frac{dx}{dt} = \epsilon - x $, where $v_t(x) = \frac{dx}{dt}$ represents the target flow vector or velocity field that the neural network $f$ aims to learn. Therefore, the core objective of Flow Matching training is to have the model-parameterized vector field $u(x,t)$ approximate the target velocity field $v_t(x)$ at time 
$t$. Accordingly, the loss is formulated as:
\begin{align}
	L_{fm}=E_{t\sim U(0,1)}[\left | \left | u(x,t) - v_t(x) \right |  \right | ^2 ]
\label{eq:flow_matching_loss}
\end{align}

To enhance training efficiency and convergence in conditional video generation tasks, we follow a unified training paradigm \cite{seedance,Hunyuanvideo,Cogvideox}. The original video is first encoded using a 3D-VAE, producing a compressed latent visual representation $x$. In the conditional generation setting, the input prompt is encoded via T5 \cite{t5}, CLIP \cite{clip}, or a vision-language model (VLM) \cite{qwen} to obtain its semantic representation $c$. A noisy version $x_t$ is then generated at a random timestep $t$ and, together with the prompt representation $c$, fed into the diffusion model to produce $u(x, t, c)$, which represents the predicted velocity field under the current condition. Within the model, conditional guidance is achieved by computing attention between the visual representation $x_t$ and the prompt $c$. Specifically, three separate projection matrices $W_q$, $W_k$, and $W_v$ are constructed to compute $Q=W_q(x_t)$, $K=W_k([x_t, c])$, $V=W_v([x_t, c])$. These are used to calculate the attention weights $A_w$ and the corresponding correlation representation $A_{out}$.
\begin{align}
    A_w = softmax(\frac{QK^T}{\sqrt{D}}),  A_{out} = A_w \times  V  
    \label{eq:attention}
\end{align}

\subsection{Distribution-Aware Rectification and Spatial Ensemble}
\subsubsection{Overview}
To address the issue that text-guided conditional diffusion models tend to overfit low-semantic tokens during training, which in turn leads to low-quality video generation at inference time, we propose Distribution-Rectification Classifier-Free Guidance (DR-CFG) and Spatial Representation Alignment (SRA) methods. These two methods tackle the problem from the perspectives of distribution correction and spatial representation alignment, respectively.

Specifically, we follow the standard training paradigm of conditional diffusion models and compute the flow loss $L_{fm}$ under the current text condition using the Flow Matching loss defined in Eq. \ref{eq:flow_matching_loss}. In addition, we feed the original noise-free data $x$ into the conditional diffusion model and extract the text–visual attention weights $A_{w|t=0}$. After applying morphological operations to $A_{w|t=0}$, we obtain a mask $A_{m|t=0}$ that represents the high-attention spatial regions corresponding to each token. Based on $A_{w|t=0}$ and $A_{m|t=0}$ for each token, we compute the token-specific loss as $L_{c_i} = \frac{\sum A_{wi|t=0} \cdot A_{mi|t=0} \cdot L_{fm}}{F \cdot H \cdot W \cdot D}$. Here, $F$ denotes the number of frames, $H$ and $W$ represent the image height and width, and $D$ is the channel dimension of the hidden features. Additionally, based on the overall textual condition description $c$, we store the value $L_{c_i}$ of each token together with its occurrence count $N_{c_i}$. 

Second, in the Distribution-Rectification Classifier-Free Guidance (DR-CFG) method, we leverage the aforementioned information to dynamically adjust the flow-matching strength according to the importance weights of tokens, thereby rectifying the model’s predicted distribution and preventing the model from overfitting to tokens with low semantic-importance density. Implementation details are provided in Section \ref{sec:dr-cfg}.

Finally, in the Spatial Representation Alignment (SRA) method, we dynamically regulate the distribution of attention weights based on token importance weights. This design prevents tokens with low-density semantic information from receiving excessive attention, which could otherwise obscure tokens containing high-density semantic information. As a result, the model’s attention allocation and representation learning for high-density semantic tokens are enhanced. Detailed implementation procedures are described in Section \ref{sec:sra}.

\begin{figure*}[t!]
\centering

\begin{subfigure}[t]{0.51\linewidth}
    \centering
    \includegraphics[width=\linewidth]{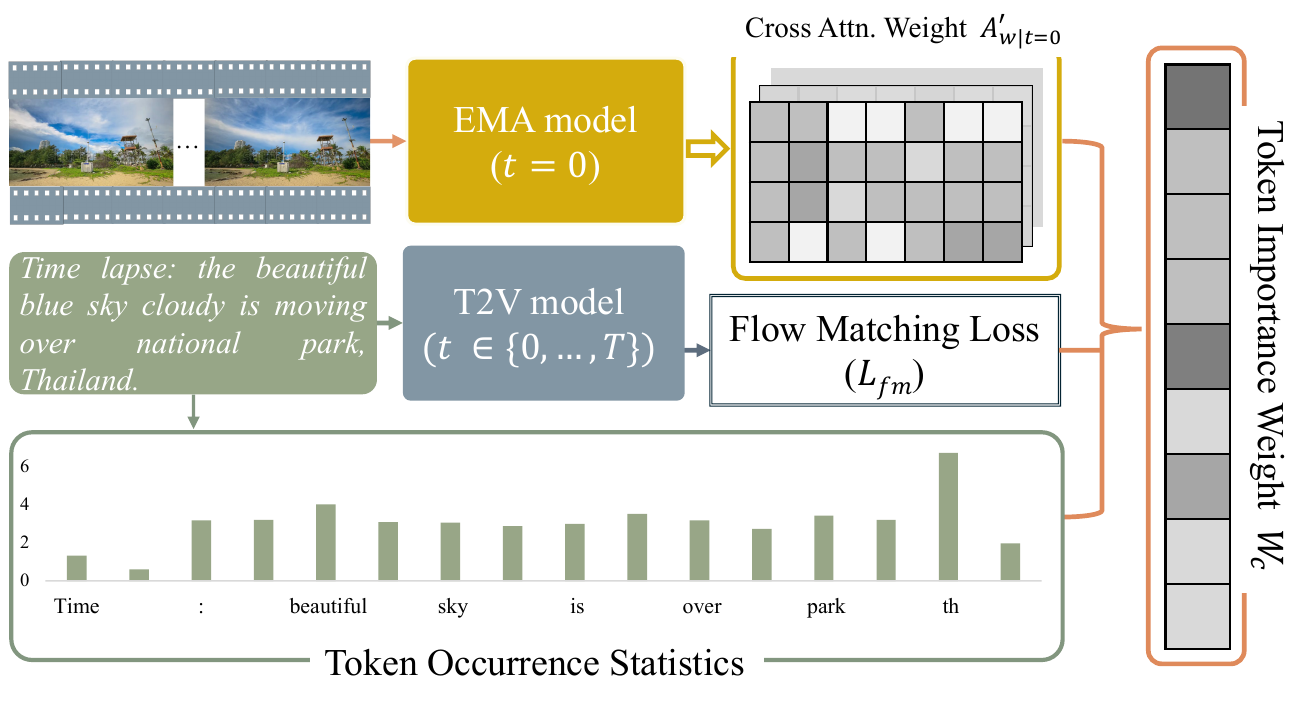}
    \caption{Computation of Importance Weights for Each Token in the Prompt.}
    \label{fig:method_overall}
\end{subfigure}
\begin{subfigure}[t]{0.48\linewidth}
    \centering
    \includegraphics[width=\linewidth]{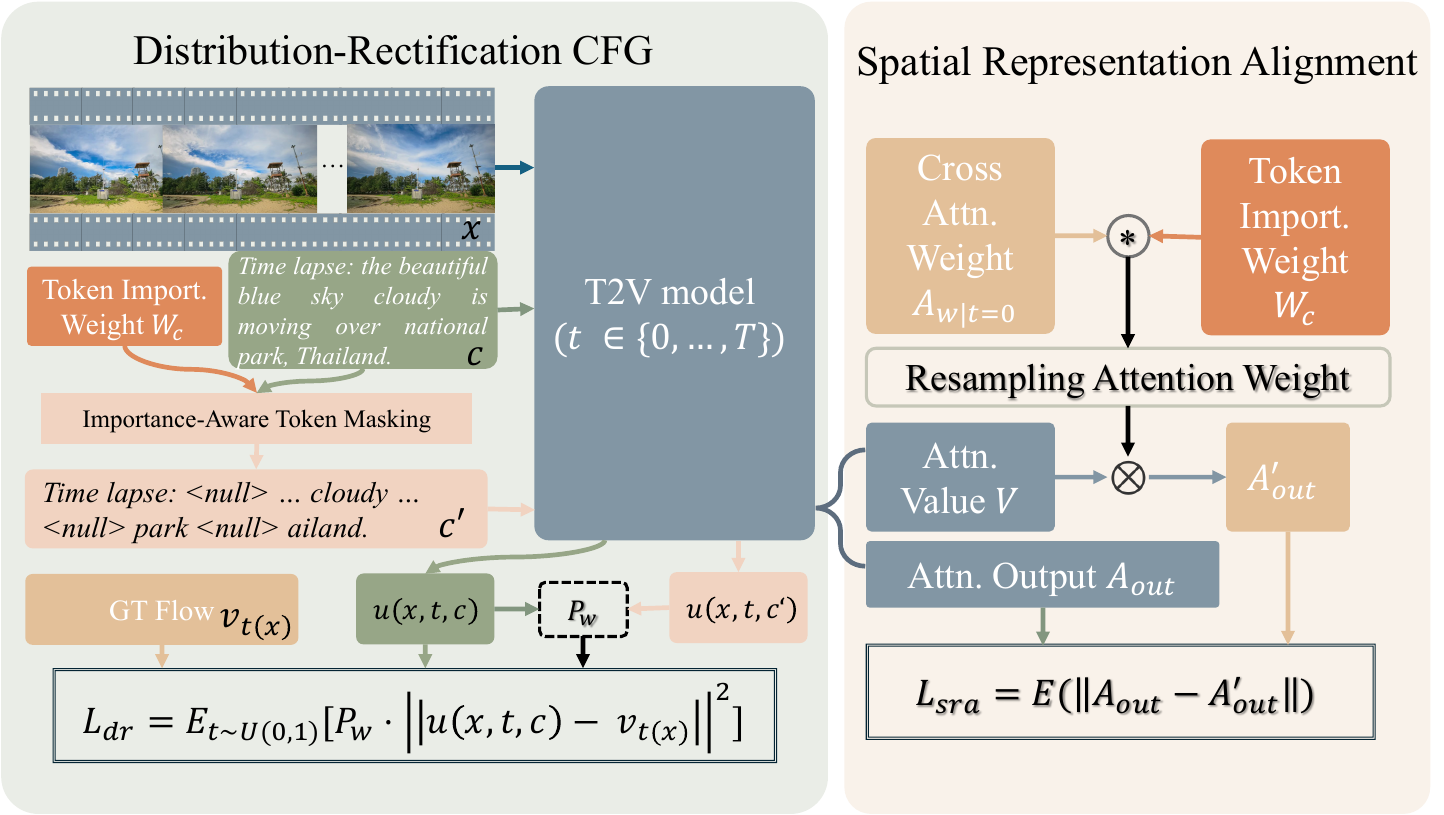}
    \caption{Computation of Distribution-Rectification Classifier-Free Guidance (DR-CFG) and Spatial Representation Alignment (SRA).}
    \label{fig:dr_cfg_sra_compute}
\end{subfigure}
\caption{Overall Computational Architecture of the Distribution-Aware Rectification and Spatial Ensemble (DARE) Method.
By computing token importance weights online and incorporating them into the subsequent DR-CFG and SRA modules, DARE corrects the distributional bias in the model’s fitting to prompt tokens.}
\label{fig:dare}
\end{figure*}

\subsubsection{Distribution-Rectification Classifier-Free Guidance}
\label{sec:dr-cfg}

To address the issue of biased distribution predictions in conditional diffusion models—caused by the model's tendency to overfit the semantic information of simple tokens during training—we propose a Distribution Rectification approach based on Classifier-Free Guidance (CFG) \cite{cfg}. 
Unlike related approaches \cite{nag}, we abandon the conventional practice of computing velocity field differences under full-token guidance. Instead, we selectively use a subset of tokens as semantic guidance based on their semantic importance to compute the velocity field. We then perform a spatially fine-grained, weighted difference with the velocity field obtained under full-token guidance, which is used to rectify the model’s distribution prediction under full-token conditions.
Specifically, we first calculate the importance weight ($W_{c_i}$) for each semantic token based on its cumulative loss ($L_{c_i}$) and frequency of occurrence ($N_{c_i}$) recorded during training, as defined in Eq. \ref{eq:sample_weight}.
\begin{align}
    W_{c_i}=\frac{L_{c_i}}{N_{c_i} \cdot \ln_{}{(1+N_{c_i})}}
    \label{eq:sample_weight}
\end{align}
% Next, we rank all tokens within the conditional text based on their weights $W_{c_i}$. A subset of conditions, denoted as $c^{'}$, is constructed by retaining a specific proportion of tokens with the lowest importance scores.
% Finally, a subset $c'$ of the conditional text $c$, consisting of low-importance weights (simple tokens with low semantic information), is used as a new conditional input. 
Finally, we sort all conditional text tokens $c$ according to $W_{c_i}$ and construct a subset $c^{'}$. This subset consists of tokens with low importance weights that correspond to low-density semantic information, while the remaining tokens are replaced with null embeddings($<\phi>$) and used as the new conditional input.
The model performs a forward pass to obtain the velocity field $u(x, t, c^{'})$, which represents the model's fit to this low-semantic conditional information. We then calculate $P_w$ as the pixel-wise difference between the low-semantic-guided velocity field $u(x, t, c^{'})$ and the ground-truth velocity field $v_t(x)$. This $P_w$ is used to reweight the loss $L_{fm}$ between the fully conditioned velocity field $u(x, t, c)$ and the ground truth $v_t(x)$. Specifically, $P_w$ reflects the model's convergence state regarding the low-importance tokens $c'$; a smaller $P_w$ indicates a higher degree of fitting to $c'$. By adjusting the weight of $L_{fm}$ via $P_w$, we amplify the contribution of high-importance tokens to the fitting process, thereby biasing the model's predicted distribution toward the influence of high-semantic information.
\begin{equation}
\begin{cases}
      P_w=\gimel ( v_t(x) - u(x, t, c^{'}) )^{2}, \\
      P_w=\frac{1}{1+e^(-P_w)}*2 -1, \\
      L_{dr}=E_{t \sim U(0,1)}[P_w \cdot ||u(x, t, c) - v_t(x)||^2]
\end{cases}
\label{eq:dr_cfg_rw}
\end{equation}
In the above expression, $\frac{1}{1+e^(-P_w)}$ designed to map the weights of $P_w$ onto the interval $[0.5, 1]$. $\gimel$ denotes the truncated gradient, where backpropagation is not performed.

\begin{table*}[ht!]
\centering
\resizebox{\textwidth}{!}{
\begin{tabular}{cccccccc}
\hline
Method & \begin{tabular}[c]{@{}c@{}}Appearance\\ Style(\%)\end{tabular}  & \begin{tabular}[c]{@{}c@{}}Overall\\ Consistency(\%)\end{tabular}  & \begin{tabular}[c]{@{}c@{}}Temporal\\ Style(\%)\end{tabular}  & \begin{tabular}[c]{@{}c@{}}Dynamic\\ Degree(\%)\end{tabular}  & \begin{tabular}[c]{@{}c@{}}Aesthetic\\ Quality(\%)\end{tabular}  & \begin{tabular}[c]{@{}c@{}}Background\\ Consistency(\%)\end{tabular}  & Scene(\%)  \\ \hline
LaVie$\S$ \cite{Lavie}  & 23.56 &  26.41  & 25.93 & 49.72 & 54.94 & 97.47 &  52.69      \\
ModelScope$\S$ \cite{Modelscope}  & 23.39 &  25.67  & 25.37 & 66.39 & 52.06 & 95.29 &  39.26      \\
VideoCrafter$\S$ \cite{VideoCrafter} & 21.57 &  25.21  & 25.42 & 89.72 & 44.41 & 92.88 &  43.36      \\ 
CogVideo$\S$ \cite{Cogvideo} & 22.01 &  7.70  & 7.80 & 42.22 & 38.18 & 95.42 &  28.24      \\ \hline

Wan 2.1 \cite{wan} & 20.80 &  20.79  & 19.93 & 88.89 & 46.67 \gain{2.55} & 96.69 &  18.02 \gain{9.08}      \\
\rowcolor{IceBlue!80}\textbf{w. DARE (Ours)}  & 21.09 \gain{0.29} & 22.76 \gain{2.00}   &  22.29 \gain{2.36}    & 95.84 \gain{6.95} & 44.12 & 97.54 \gain{0.85} &  8.94     \\ \hline
Seedance1.0\_exp \cite{seedance}  & 22.37  & 27.37         & 25.27       & 80.55  & 58.45        & 96.37         & 47.53   \\
\rowcolor{IceBlue!80}\textbf{w. DARE (Ours)}  & 22.66 \gain{0.29}  &  27.71 \gain{0.34}  & 25.39 \gain{0.12} & 87.50 \gain{6.95} & 58.63 \gain{0.18} &  96.65 \gain{0.28} & 51.82 \gain{4.29} \\ \hline
\diagbox{}{}    & \begin{tabular}[c]{@{}c@{}}Motion\\ Smoothness(\%)\end{tabular}  & \begin{tabular}[c]{@{}c@{}}Imaging\\ Quality(\%)\end{tabular}  & \begin{tabular}[c]{@{}c@{}}Object\\ Class(\%)\end{tabular}   & \begin{tabular}[c]{@{}c@{}}Spatial\\ Relationship(\%)\end{tabular}  & Color  & \begin{tabular}[c]{@{}c@{}}Subject\\ Consistency(\%)\end{tabular}  & \begin{tabular}[c]{@{}c@{}}Multiple\\ Objects(\%)\end{tabular}  \\ \hline

LaVie$\S$ \cite{Lavie}   & 96.38 &  61.90  & 91.82 & 34.09 & 86.39 & 91.41 &  33.32      \\
ModelScope$\S$ \cite{Modelscope}  & 95.79 &  58.57  & 82.25 & 33.68 & 81.72 & 89.87 &  38.98      \\
VideoCrafter$\S$ \cite{VideoCrafter} & 91.79 &  57.22  & 87.34 & 36.74 & 78.84 & 86.24 &  25.93      \\ 
CogVideo$\S$ \cite{Cogvideo}  & 96.47 &  41.03  & 73.40 & 18.24 & 79.57 & 92.19 &  18.11      \\ \hline

Wan 2.1 \cite{wan} & 98.46 & 49.22 & 54.43 & 49.16 & 84.79 \gain{1.53} & 91.69 & 25.60 \\
\rowcolor{IceBlue!80}\textbf{w. DARE (Ours)}  & 99.36 \gain{0.9} &  59.04 \gain{9.82} &   54.67 \gain{0.24}  & 55.95 \gain{6.79} & 83.26 &  93.14 \gain{1.45}   & 35.52 \gain{9.92}  \\ \hline
Seedance1.0\_exp \cite{seedance} & 98.39  & 67.07    & 87.26   & 49.26    & 84.69    & 95.76     & 76.45    \\
\rowcolor{IceBlue!80}\textbf{w. DARE (Ours)} &  98.39 \gain{0.0}  & 67.96 \gain{0.89} & 92.56 \gain{5.3}  & 54.41 \gain{5.15}  & 87.39 \gain{2.7} & 95.88 \gain{0.12} & 79.65 \gain{3.2} \ \\ \hline
\end{tabular}}
\caption{Evaluation results of generated outputs under the VBench \cite{vbench,Vbench++} frameworks using the prompts provided by VBench. $\S$ indicates that the results of the corresponding methods are taken from \cite{vbench}, while all other results are obtained through inference and evaluation conducted by us under the same experimental environment.}
\label{tab:compare_sota_in_vbench}
\end{table*}

\subsubsection{Spatial Representation Alignment}
\label{sec:sra}
To address the dominance of low-semantic-density tokens in attention, which often leads the model to overlook the spatial and distributional guidance of semantically important tokens, we introduce Spatial Representation Alignment (SRA). Unlike methods that focus on alignment in the feature-level representation space \cite{repa}, we adopt a self-supervised training paradigm. By analyzing text-conditioned spatial attention weights and attention representations, we increase the influence of tokens carrying complex semantic information, thereby strengthening the weight distribution and guidance intensity of relevant tokens. This approach enables targeted enhancement of the spatial representations of complex semantic tokens.
Specifically, we employ a self-supervised attention representation constraint with the timestep set to $t=0$. The noise-free visual representations, together with the complete conditional text, are fed into the diffusion model for a single forward pass, from which the resulting attention weights $A_{w|t=0}$ are extracted.
Since $A_{w|t=0}$ directly reflects the visual regions focused on by different text tokens, we apply token importance weights $W_c$ to $A_{w|t=0}$ to adjust the attention intensity. This allows for the precise correction of focus strength across various tokens.
\begin{align}
      A_{w|t=0}^{'}=A_{w|t=0} \cdot (\frac{W_{c}}{\sigma(W_{c})})   
\label{eq:attn_w_rw}
\end{align}
In the equation above, the tensor $A_{w|t=0}$ has dimensions of $(B, {head}, F \times H \times W, K)$, where $K$ denotes the number of tokens in the current conditional text (i.e., the length of set $c$), $B$ represents the number of samples, and $head$ represents the number of attention heads. The weight vector $W_c$ initially has a dimension of $(K,)$. We amplify the variance between high and low token importance weights by scaling the weights by $\frac{W_{c}}{\sigma(W_{c})}$, where $\sigma(W_{c})$ represents the standard deviation of the weight distribution. During computation, $W_c$ is reshaped or broadcasted to $(1, 1, 1, K)$ to re-weight the attention scores based on individual token importance. Consequently, this operation modifies the spatial weight coefficients for different tokens within $A_{w|t=0}$.

Finally, considering that the values of the attention weights are constrained within the range $[0, 1]$, applying direct supervision to these weights often results in vanishingly small gradients, which fails to provide an effective spatial constraint. To address this, we instead utilize the attention Value $V$ at timestep $t$ (as defined in Eq. \ref{eq:attention}) and compute the weighted product with the attention weight $A_{w|t=0}^{'}$. This process yields $A_{out}^{'}$, an attention feature that highlights regions represented by high-importance tokens. We then use $A_{out}^{'}$ to enforce the supervisory constraints.
% \begin{equation}
% \begin{cases}
%       A_{out}^{'} = \gimel (A_{w|t=0}^{'} \times V), \\
%       L_{sra}=E[||A_{out}^{'} - A_{out}||^2]
% \end{cases}
% \label{eq:sra_loss}
% \end{equation}
\begin{align}
A_{out}^{'} = \gimel (A_{w|t=0}^{'} \times V), L_{sra}=E[||A_{out}^{'} - A_{out}||^2]
\label{eq:sra_loss}
\end{align}

\subsubsection{Training Paradigm}
\label{sec:implement_detail}
We observe that jointly introducing $L_{sra}$ and $L_{dr}$ often leads to training instability and model collapse. We attribute this issue primarily to the strong representation-level constraints imposed by $L_{sra}$. In contrast, the $L_{dr}$ module, built upon Flow Matching, is easily dominated by the gradients introduced by these strong constraints. As a result, its gradient contributions are overwhelmed, which can lead to instability and even training collapse in the middle and later stages. In the Appendix (Section Implementation and Training Details), we provide a detailed analysis of this issue and propose a simple yet effective solution. Additional implementation and inference details are also included in the Appendix.

% 这一段的叙述逻辑感觉是实验性质，我的建议是这里上来就是为了解决349-352行的这个问题，你设计这么一个training scheme，然后这个小点你甚至可以写到你introduction里为了整个工作描述的完整性。然后你当前这部分的实验发现，也就是不用你的调节机制的话training会不稳定这个事，放在ablation study里面，就多一个小实验单独ablate一下你的training scheme。
% 另外图3建议不要那样叠放，把曲线挡住了看不清

\section{Experiments}

% [TODO]: 和intro 中提到的一致，加入attention weight，对比DARE前后，Attention Map变化。==> 实验部分最终是想验证这个工作大idea是不是立得住，这里我还是建议可视化用方法之前和用方法之后的tam来个直接证明，因为我们抛出的核心问题是从推理过程可解释性角度出发的，现有实验数据和图4都只是现象上对得上，但不直接。

\subsection{Datasets and Evaluation Metrics}
We train our model on the WebVid \cite{webvid} dataset, which contains over 10 million video–text pairs covering a wide range of application scenarios and exhibiting substantial semantic diversity. For evaluation, we additionally employ two sets of textual prompts to guide video generation: one constructed with the assistance of ChatGPT and the other provided by VBench \cite{vbench,Vbench++}.

To assess model performance, we generate videos conditioned on these two prompt sets and evaluate the results using the EvalCrafter \cite{Evalcrafter} and VBench \cite{vbench,Vbench++} evaluation frameworks. EvalCrafter primarily focuses on metrics such as temporal consistency and visual quality, while VBench further extends the evaluation dimensions to include aesthetic quality, object consistency, and semantic alignment. A more detailed description of the experimental setup and evaluation protocol is provided in the Appendix.

\subsection{Comparison and Analysis with state-of-the-art Methods}

To comprehensively validate the effectiveness of our method, we conduct experiments on both the open-source model Wan 2.1 \footnote{Wan 2.1 is a 14B-parameter model, with its pretrained weights available at: \url{https://huggingface.co/Wan-AI/Wan2.1-T2V-14B}.} \cite{wan} and the closed-source model Seedance1.0\_exp \cite{seedance} \footnote{Seedance1.0\_exp is a base model built upon Seedance 1.0, primarily used for internal experiments, with a total parameter count of 7B. To ensure a fair comparison, we fine-tune both models based on their respective pretrained weights.}. To ensure a fair comparison, we fine-tune the base models under the same environment and dataset settings.
First, we perform inference evaluation using the prompts provided by the VBench benchmark \cite{vbench} \footnote{Prompt Link: \url{https://github.com/Vchitect/VBench/blob/master/vbench/VBench_full_info.json}, Evaluation code Link: \url{https://github.com/Vchitect/VBench/tree/master/vbench}}. As shown in Table \ref{tab:compare_sota_in_vbench}, the models are quantitatively evaluated from 14 different dimensions. The results clearly demonstrate that introducing DARE during training effectively improves the inference performance of the base models. For example, on semantic consistency metrics—Object Class, Multiple Objects, and Spatial Relationship—our method improves performance over Seedance1.0\_exp by 5.3\%, 3.2\%, and 5.15\%, respectively, and over Wan 2.1 by 0.24\%, 9.92\%, and 6.79\%. Moreover, notable improvements are also observed on metrics related to aesthetic quality and motion dynamics.
We attribute these improvements primarily to two factors. First, during training, dynamically adjusting the contribution of different tokens according to the model’s learning status helps the model better capture the distributional guidance of high semantic-density tokens, thereby improving the semantic consistency between the generated content and the textual description. Second, spatial representation alignment effectively mitigates spatiotemporal collapse and enhances dynamic motion quality.

In addition, we conduct a comprehensive evaluation of the generated results using the EvalCrafter benchmark \cite{Evalcrafter}; detailed results are provided in the Experiments section of the supplementary. Since the prompts in VBench are relatively simple, we further construct an additional set of 100 prompts using GPT, considering aspects such as semantic complexity, animation, and visual effects. These prompts are used to evaluate model performance under more complex conditions. The generated results are assessed using both VBench and EvalCrafter, and the detailed evaluation results are reported in the supplementary.

\subsection{Ablation Studies}
\begin{table*}[h!]
\centering
\resizebox{\textwidth}{!}{
\begin{tabular}{cccccccc}
\hline
Method                                                     & \begin{tabular}[c]{@{}c@{}}Appearance\\ Style(\%)\end{tabular}  & \begin{tabular}[c]{@{}c@{}}Overall\\ Consistency(\%)\end{tabular} & \begin{tabular}[c]{@{}c@{}}Temporal\\ Style(\%)\end{tabular} & \begin{tabular}[c]{@{}c@{}}Dynamic\\ Degree(\%)\end{tabular}       & \begin{tabular}[c]{@{}c@{}}Aesthetic\\ Quality(\%)\end{tabular} & \begin{tabular}[c]{@{}c@{}}Background\\ Consistency(\%)\end{tabular} & Scene(\%)                                                      \\ \hline
w. - & 22.37 & 27.37 & 25.27 & 80.55 & 58.45 & 96.37 & 47.53 \\
w. SRA & 22.30 & 22.88 & 20.92 & 23.61 & 47.07 & 81.44 & 14.46 \\
w. DR-CFG & 22.38 & 27.62 & 24.91 & 81.95 & 58.57 & \textbf{96.87} & 46.80 \\
\rowcolor{LightGray} \begin{tabular}[c]{@{}c@{}}w. DARE\\ w. 15\%\end{tabular} & \textbf{22.66} & \textbf{27.71} & 25.39 & \textbf{87.50} & 58.63 & 96.65 & \textbf{51.82} \\
w. 20\% & 22.65 & 27.83 & \textbf{25.48} & 77.78 & \textbf{58.81} & 96.72 & 49.86 \\
w. 25\% & 22.24 & 27.78 & 25.31 & 79.17 & 58.45 & 96.63 & 49.64 \\ \hline
\diagbox{}{}    & \begin{tabular}[c]{@{}c@{}}Motion\\ Smoothness(\%)\end{tabular} & \begin{tabular}[c]{@{}c@{}}Imaging\\ Quality(\%)\end{tabular}     & \begin{tabular}[c]{@{}c@{}}Object\\ Class(\%)\end{tabular}   & \begin{tabular}[c]{@{}c@{}}Spatial\\ Relationship(\%)\end{tabular} & Color(\%)                                                       & \begin{tabular}[c]{@{}c@{}}Subject\\ Consistency(\%)\end{tabular}    & \begin{tabular}[c]{@{}c@{}}Multiple\\ Objects(\%)\end{tabular} \\ \hline
w. - & 98.39 & 67.07 & 87.26 & 49.26 & 84.69 & 95.76 & 76.45 \\
w. SRA & 97.91 & 60.78 & 53.09 & 42.04 & 87.74 & 73.42 & 19.89 \\
w. DR-CFG & 98.28 & 66.86 & 91.46 & 49.11 & \textbf{90.30} & 95.30 & \textbf{76.83} \\
\rowcolor{LightGray} \begin{tabular}[c]{@{}c@{}}w. DARE\\ w. 15\%\end{tabular} & \textbf{98.39} & \textbf{67.96} & \textbf{92.56} & \textbf{54.41} & 87.39 & \textbf{95.88} & 79.65 \\
w. 20\% & 98.26 & 67.53 & 87.18 & 51.51 & 88.65 & 95.49 & 79.27 \\
w. 25\% & 98.37 & 67.37 & 92.33 & 52.74 & 84.64 & 95.26 & 74.16 \\ \hline
\end{tabular}}
\caption{Ablation results analyzing the effects of Distribution-Rectification Classifier-Free Guidance (DR-CFG), Spatial Representation Alignment (SRA), and varying proportions of low-semantic tokens on model performance.}
\label{tab:ablation}
\end{table*}

We conduct ablation studies under the VBench evaluation framework. For each ablated component, inference is performed using the corresponding evaluation subsets provided by VBench, and all metrics are computed accordingly. The evaluation results are reported in Table \ref{tab:ablation}.

\subsubsection{Ablation Study of DR-CFG and SRA}
We separately analyze Distribution-Rectification Classifier-Free Guidance (DR-CFG) and Spatial Representation Alignment (SRA), as shown in the first four rows of Table \ref{tab:ablation}. As discussed in Section \ref{sec:implement_detail}, using SRA alone tends to cause the model to overfit semantic information at the token level while neglecting spatial–temporal consistency. This results in severe inter-frame discontinuities and visual artifacts. Consequently, as shown in the second row of Table \ref{tab:ablation}, employing only SRA leads to varying degrees of performance degradation across evaluation perspectives. In contrast, compared with the third and fourth rows, using DR-CFG alone yields noticeable improvements in overall generation quality and exhibits more stable behavior. Notably, when DR-CFG and SRA are jointly applied, we observe a significant performance gain. This is primarily because SRA effectively accelerates model learning and convergence, while DR-CFG regulates the model’s fitting behavior across different tokens, enabling superior generation quality and evaluation performance under the same training conditions.

\subsubsection{Ablation Study on the Selection Ratio of Low-Semantic Tokens}
During training, we dynamically update the importance weights of different tokens and perform targeted selection accordingly. Considering real-world scenarios where text prompts are often short, setting an excessively high selection ratio may discard semantically informative tokens, leading to underfitting. To address this issue, we conduct an ablation study on the proportion of low-semantic tokens and evaluate ratios of 15\%, 20\%, and 25\%, as reported in Rows 4–6 of Table \ref{tab:ablation}.
Under a unified evaluation protocol, we observe that when the proportion of low-semantic tokens is 15\%, the average score reaches 67.61\%. In comparison, the average scores drop to 66.22\% and 66.01\% for ratios of 20\% and 25\%, respectively. These results indicate that selecting 15\% low-semantic tokens yields relatively better performance gains across evaluation metrics.

\begin{figure}[h]
\centering
    \includegraphics[width=0.48\textwidth]{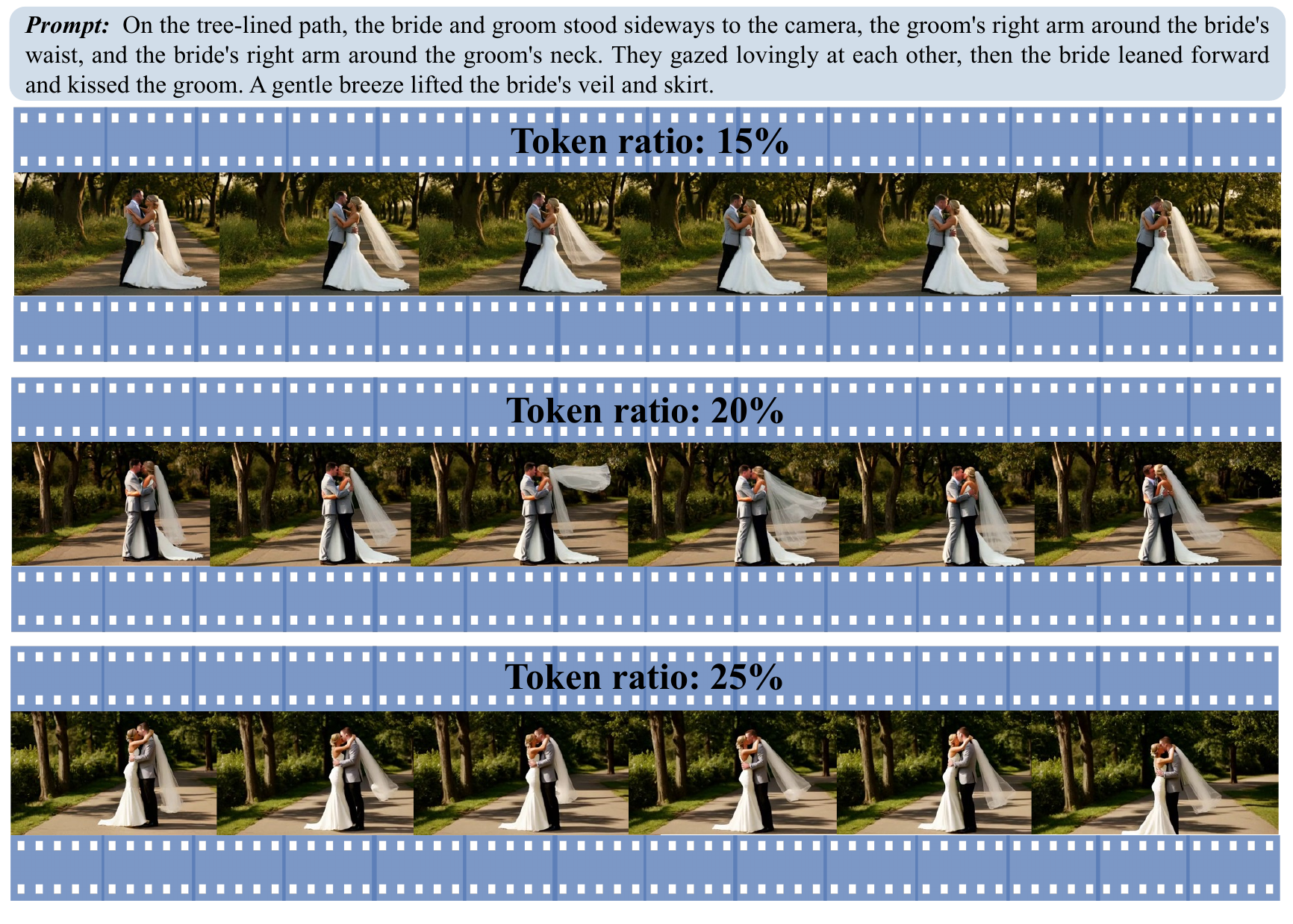} 
  \caption{Visual evaluation of the effects of varying proportions of low-semantic tokens on model performance.}
    \label{fig:vis_ablation}
\end{figure}

Furthermore, to provide an intuitive comparison of how different ratios affect the model’s distribution and generation capability, we visualize the generation results under the three settings in Fig.~\ref{fig:vis_ablation}. As shown in Fig.~\ref{fig:vis_ablation}, when the low-semantic token ratio is set to 15\%, applying differential reweighting constraints to the model-predicted flow matching loss leads to clear advantages in terms of semantic consistency with the conditioning input and spatial structure alignment in the generated content.

\subsection{Visualization Analysis of Distribution and Attention Rectification Effects}
To provide a more comprehensive analysis of the practical gains introduced by our method, we examine the guidance and contribution of different types of tokens during inference. Specifically, we extract token-level signals and evaluate the model’s behavioral changes before and after introducing DARE from the perspective of attention scores.

\begin{figure*}[h!]
\centering
% \vspace{-10pt}
\begin{subfigure}[t]{0.48\linewidth}
    \centering
    \includegraphics[width=\linewidth]{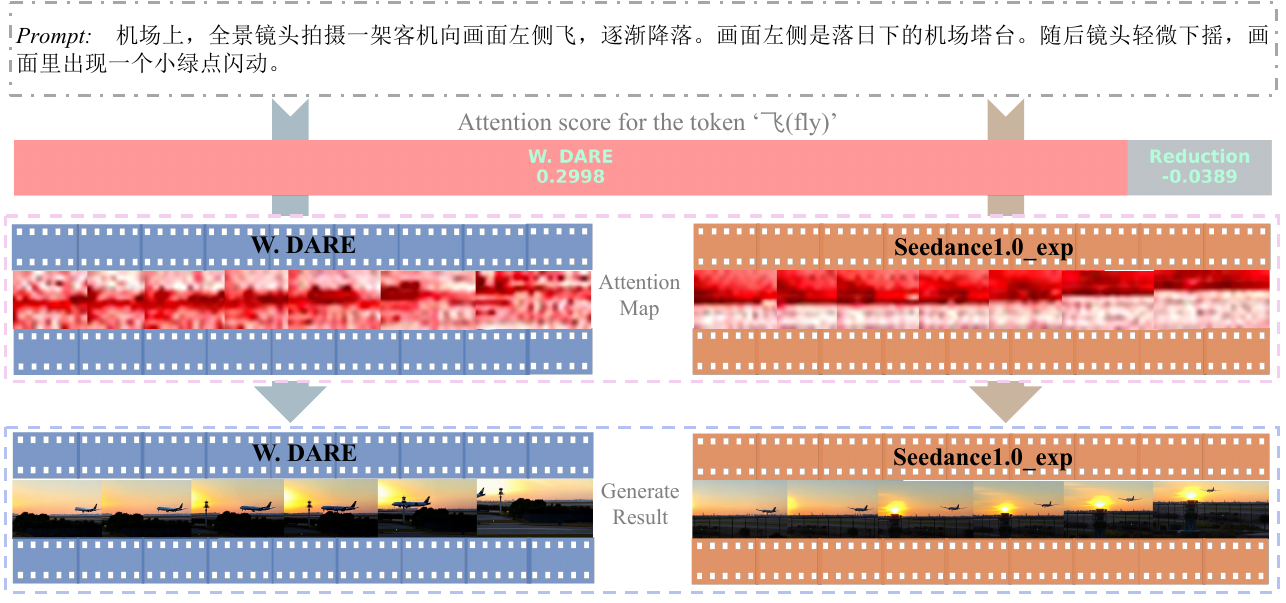}
    \caption{Impact of Low-Semantic-Density (High-Frequency) Tokens on Attention Scores and Distributions During Inference.}
    \label{fig:quan_ana_1}
\end{subfigure}
\begin{subfigure}[t]{0.51\linewidth}
    \centering
    \includegraphics[width=\linewidth]{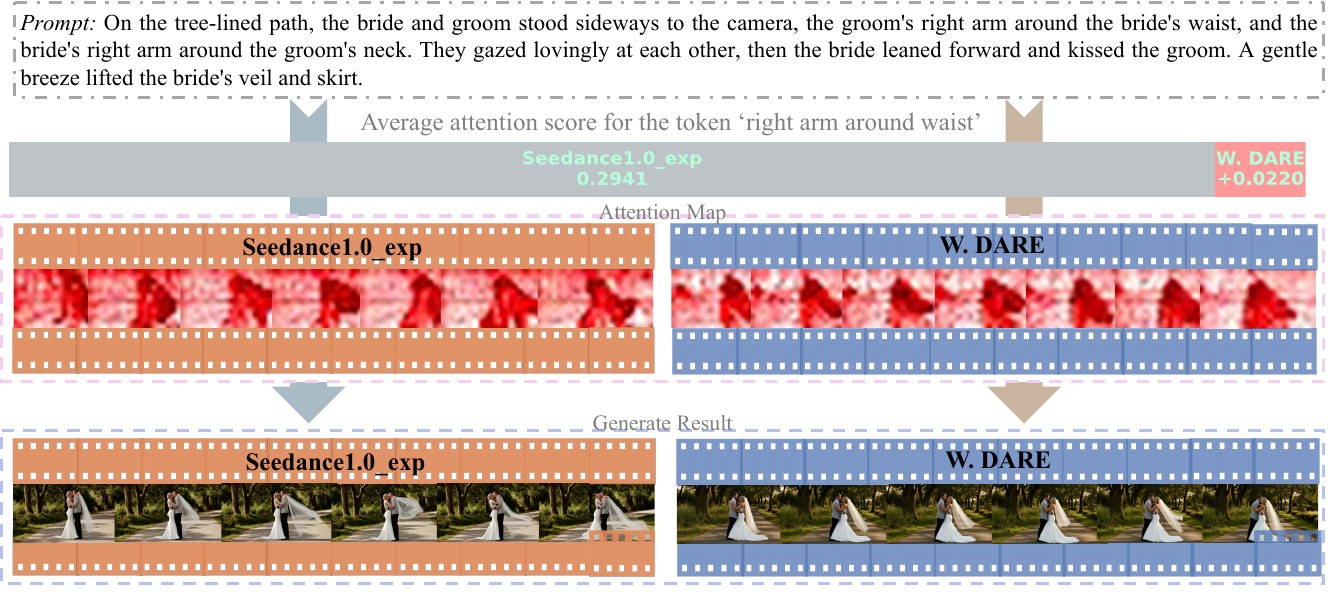}
    \caption{Impact of High-Semantic-Density (Low-Frequency) Tokens on Attention Scores and Distributions During Inference.}
    \label{fig:quan_ana_2}
\end{subfigure}
\caption{Visualization of Token-Level Guidance Dynamics Before and After DARE from the Perspective of Attention Scores and Distributions.}
\label{fig:quan_ana}
\end{figure*}

As illustrated in Fig.~\ref{fig:quan_ana_1}, for the low semantic-density (high-frequency) token “fly”, the attention score decreases after integrating DARE into Seedance1.0\_exp. In the baseline model, this token produces attention that broadly covers the image content, causing the model to be excessively influenced by “fly” during inference and consequently overlook tokens related to “gradually landing”.

Similarly, in Fig.~\ref{fig:quan_ana_2}, for the high semantic-density token group “right arm around waist”, the average attention score increases after introducing DARE. The attention map further shows that, in the baseline model, the attention associated with this token group does not converge and exhibits a clear uncertainty distribution. This unstable attention allocation leads to semantic confusion during inference and results in incorrect spatial constraints in the generated outputs.

From both the quantitative attention scores and the qualitative attention maps, it is evident that a severely imbalanced token distribution causes the model to overfit low semantic-density tokens, while high semantic-density tokens are insufficiently learned. As a result, during inference, low semantic-density tokens exert disproportionately strong guidance, leading to incorrect content generation. By incorporating DARE during training, we dynamically correct the model’s distributional bias and recalibrate token-level attention intensity. This effectively mitigates the imbalance issue and significantly improves generation quality. Additionally, we include in the appendix an evaluation of the attention scores for tail tokens containing high-density semantic information. The results provide an intuitive illustration that the introduction of DARE effectively enhances the model’s ability to understand and learn from high-density semantic information.

\begin{figure*}[h!]
    \centering
    \includegraphics[width=\linewidth]{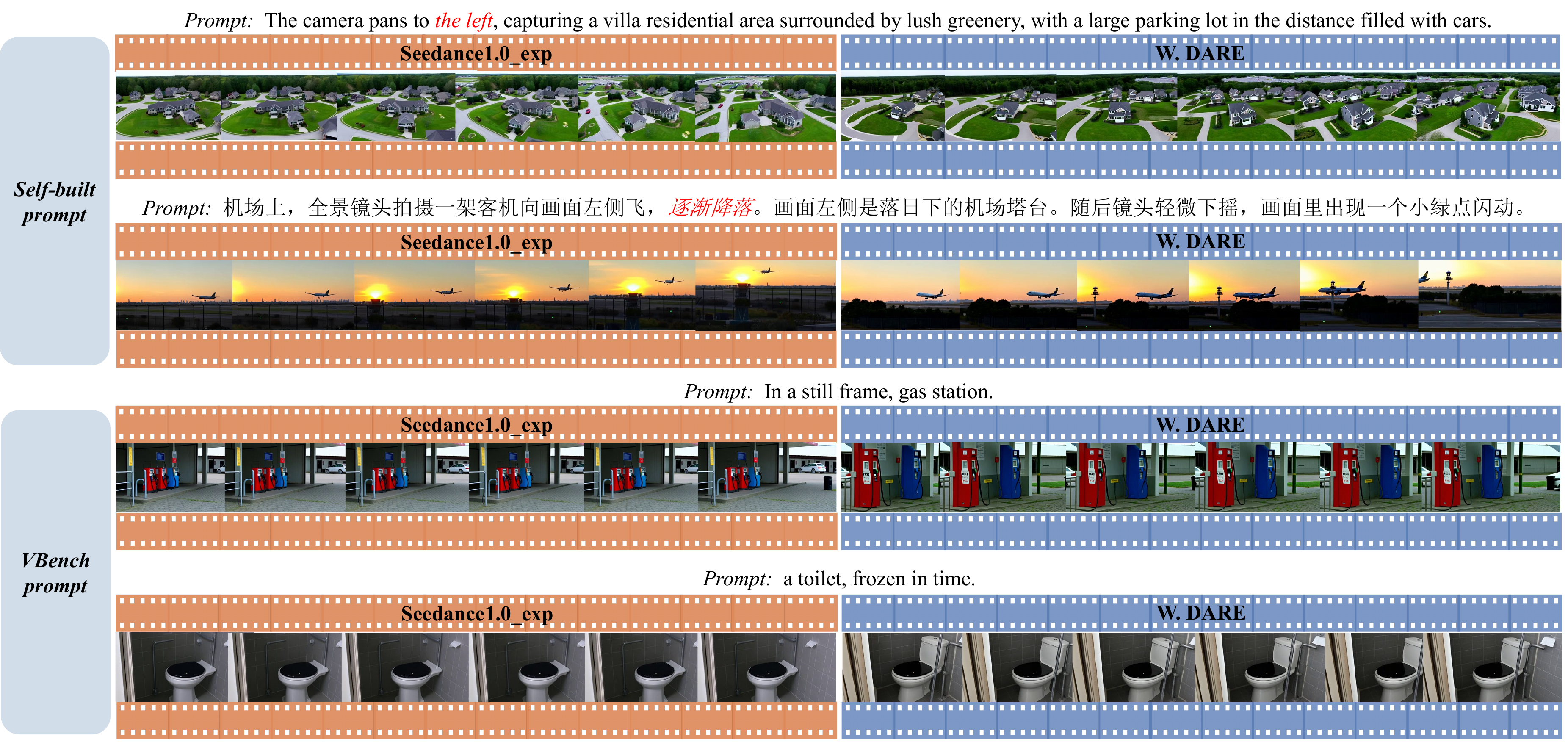}
    \caption{Visual quality evaluation of generation results based on a self-constructed evaluation dataset and the VBench benchmark.}
    \label{fig:compare_seedance}
\end{figure*}

\subsection{Qualitative Analysis}
We conduct video generation and quality evaluation by guiding the model with textual prompts from both our self-constructed evaluation dataset and the VBench benchmark. As shown in Fig.~\ref{fig:compare_seedance}, we compare our results with those produced by the current state-of-the-art video generation model, Seedance \cite{seedance}. On the self-constructed evaluation dataset, our textual prompts are relatively more complex, which leads to more noticeable differences in the generated results. For example, in the first row, the prompt specifies “\textit{The camera pans to the left},” whereas Seedance, following a default tendency, moves the camera to the right. In contrast, our method accurately interprets the instruction and rotates the camera to the left. Similarly, in the second row, Seedance ignores the semantic information related to the \textit{airplane landing}, resulting in deviations in the generated video.
In the VBench evaluation set, the prompts are generally simpler and shorter, so the overall video content does not exhibit significant differences. However, substantial disparities can be observed in the quality of generated entities, such as “fuel tanks” and “toilets.” Compared with Seedance, our method produces results that are more accurate and more consistent with real-world appearances. Additional qualitative comparisons of video generation results are provided in the appendix.

\section{Conclusion}
% This paper investigates a commonly overlooked issue in conditional diffusion models: semantic token imbalance. In this setting, the discretized distribution of token semantic representations, the long-tailed distribution of tokens, and inconsistencies in the cross-attention mechanism cause the model to underutilize guidance from semantically dense tokens. Instead, the model tends to bias toward tokens carrying lower information density, resulting in a semantic gap between the generated content and the intended semantic guidance. To address this problem, we propose Distribution-Aware Rectification and Spatial Ensemble (DARE), a unified framework that improves conditional diffusion from both distributional and spatial perspectives.
% Specifically, we introduce Distribution-Rectified Classifier-Free Guidance (DR-CFG) to mitigate distributional bias during training by suppressing the dominance of easily fitted tokens, enabling the model to learn a more balanced conditional distribution. In addition, we propose Spatial Representation Alignment (SRA), which enhances the spatial influence of semantically important tokens through importance-aware attention reweighting and representation consistency learning.
% Extensive experiments demonstrate that our approach consistently improves semantic fidelity and generation quality across multiple benchmarks. We hope this work highlights the importance of distribution-aware training and semantic-aware spatial alignment for advancing controllable and reliable conditional diffusion models.
This paper investigates the overlooked issue of semantic token imbalance in conditional diffusion models, where long-tailed token distributions and cross-attention inconsistencies cause semantically dense tokens to be underutilized.
To address this problem, we propose Distribution-Aware Rectification and Spatial Ensemble (DARE), a unified framework that improves conditional diffusion from both distributional and spatial perspectives. Specifically, DR-CFG mitigates distributional bias during training by suppressing the gradients of high-frequency, low-density semantic tokens. In parallel, SRA enhances the spatial influence of semantically important tokens through importance-aware attention reweighting and representation alignment.
Extensive experiments demonstrate consistent improvements in semantic fidelity and generation quality across multiple benchmarks. We believe our findings highlight the importance of distribution-aware training and semantic-aware spatial alignment for advancing reliable and controllable diffusion models. Furthermore, the appendix also discusses the limitations of the current work and outlines potential directions for future research.

\bibliography{main}

@article{elu_noise_schedule,
  title={Elucidating the design space of diffusion-based generative models},
  author={Karras, Tero and Aittala, Miika and Aila, Timo and Laine, Samuli},
  journal={Advances in neural information processing systems},
  volume={35},
  pages={26565--26577},
  year={2022}
}

@article{score_noise_schedule,
  title={Score-based generative modeling through stochastic differential equations},
  author={Song, Yang and Sohl-Dickstein, Jascha and Kingma, Diederik P and Kumar, Abhishek and Ermon, Stefano and Poole, Ben},
  journal={arXiv preprint arXiv:2011.13456},
  year={2020}
}

@article{flow_matching,
  title={Flow matching for generative modeling},
  author={Lipman, Yaron and Chen, Ricky TQ and Ben-Hamu, Heli and Nickel, Maximilian and Le, Matt},
  journal={arXiv preprint arXiv:2210.02747},
  year={2022}
}

@inproceedings{clip,
  title={Learning transferable visual models from natural language supervision},
  author={Radford, Alec and Kim, Jong Wook and Hallacy, Chris and Ramesh, Aditya and Goh, Gabriel and Agarwal, Sandhini and Sastry, Girish and Askell, Amanda and Mishkin, Pamela and Clark, Jack and others},
  booktitle={International conference on machine learning},
  pages={8748--8763},
  year={2021},
  organization={PmLR}
}

@article{t5,
  title={Exploring the limits of transfer learning with a unified text-to-text transformer},
  author={Raffel, Colin and Shazeer, Noam and Roberts, Adam and Lee, Katherine and Narang, Sharan and Matena, Michael and Zhou, Yanqi and Li, Wei and Liu, Peter J},
  journal={Journal of machine learning research},
  volume={21},
  number={140},
  pages={1--67},
  year={2020}
}

@article{qwen,
  title={Qwen Technical Report},
  author={Jinze Bai and Shuai Bai and Yunfei Chu and Zeyu Cui and Kai Dang and Xiaodong Deng and Yang Fan and Wenbin Ge and Yu Han and Fei Huang and Binyuan Hui and Luo Ji and Mei Li and Junyang Lin and Runji Lin and Dayiheng Liu and Gao Liu and Chengqiang Lu and Keming Lu and Jianxin Ma and Rui Men and Xingzhang Ren and Xuancheng Ren and Chuanqi Tan and Sinan Tan and Jianhong Tu and Peng Wang and Shijie Wang and Wei Wang and Shengguang Wu and Benfeng Xu and Jin Xu and An Yang and Hao Yang and Jian Yang and Shusheng Yang and Yang Yao and Bowen Yu and Hongyi Yuan and Zheng Yuan and Jianwei Zhang and Xingxuan Zhang and Yichang Zhang and Zhenru Zhang and Chang Zhou and Jingren Zhou and Xiaohuan Zhou and Tianhang Zhu},
  journal={arXiv preprint arXiv:2309.16609},
  year={2023}
}

@article{seedance,
  title={Seedance 1.0: Exploring the Boundaries of Video Generation Models},
  author={Gao, Yu and Guo, Haoyuan and Hoang, Tuyen and Huang, Weilin and Jiang, Lu and Kong, Fangyuan and Li, Huixia and Li, Jiashi and Li, Liang and Li, Xiaojie and others},
  journal={arXiv preprint arXiv:2506.09113},
  year={2025}
}

@article{Cogvideox,
  title={Cogvideox: Text-to-video diffusion models with an expert transformer},
  author={Yang, Zhuoyi and Teng, Jiayan and Zheng, Wendi and Ding, Ming and Huang, Shiyu and Xu, Jiazheng and Yang, Yuanming and Hong, Wenyi and Zhang, Xiaohan and Feng, Guanyu and others},
  journal={arXiv preprint arXiv:2408.06072},
  year={2024}
}

@article{Hunyuanvideo,
  title={Hunyuanvideo: A systematic framework for large video generative models},
  author={Kong, Weijie and Tian, Qi and Zhang, Zijian and Min, Rox and Dai, Zuozhuo and Zhou, Jin and Xiong, Jiangfeng and Li, Xin and Wu, Bo and Zhang, Jianwei and others},
  journal={arXiv preprint arXiv:2412.03603},
  year={2024}
}

@article{cfg,
  title={Classifier-free diffusion guidance},
  author={Ho, Jonathan and Salimans, Tim},
  journal={arXiv preprint arXiv:2207.12598},
  year={2022}
}

@article{make-a-video,
  title={Make-a-video: Text-to-video generation without text-video data},
  author={Singer, Uriel and Polyak, Adam and Hayes, Thomas and Yin, Xi and An, Jie and Zhang, Songyang and Hu, Qiyuan and Yang, Harry and Ashual, Oron and Gafni, Oran and others},
  journal={arXiv preprint arXiv:2209.14792},
  year={2022}
}

@inproceedings{pixeldance,
  title={Make pixels dance: High-dynamic video generation},
  author={Zeng, Yan and Wei, Guoqiang and Zheng, Jiani and Zou, Jiaxin and Wei, Yang and Zhang, Yuchen and Li, Hang},
  booktitle={Proceedings of the IEEE/CVF Conference on Computer Vision and Pattern Recognition},
  pages={8850--8860},
  year={2024}
}

@inproceedings{unet,
  title={U-net: Convolutional networks for biomedical image segmentation},
  author={Ronneberger, Olaf and Fischer, Philipp and Brox, Thomas},
  booktitle={International Conference on Medical image computing and computer-assisted intervention},
  pages={234--241},
  year={2015},
  organization={Springer}
}

@inproceedings{Glyph-byt5,
  title={Glyph-byt5: A customized text encoder for accurate visual text rendering},
  author={Liu, Zeyu and Liang, Weicong and Liang, Zhanhao and Luo, Chong and Li, Ji and Huang, Gao and Yuan, Yuhui},
  booktitle={European Conference on Computer Vision},
  pages={361--377},
  year={2024},
  organization={Springer}
}

@inproceedings{sd35,
  title={Scaling rectified flow transformers for high-resolution image synthesis},
  author={Esser, Patrick and Kulal, Sumith and Blattmann, Andreas and Entezari, Rahim and M{\"u}ller, Jonas and Saini, Harry and Levi, Yam and Lorenz, Dominik and Sauer, Axel and Boesel, Frederic and others},
  booktitle={Forty-first international conference on machine learning},
  year={2024}
}

@misc{flux.1,
    author={Black Forest Labs},
    title={FLUX},
    year={2024},
    howpublished={\url{https://github.com/black-forest-labs/flux}},
}

@article{wan,
  title={Wan: Open and advanced large-scale video generative models},
  author={Wan, Team and Wang, Ang and Ai, Baole and Wen, Bin and Mao, Chaojie and Xie, Chen-Wei and Chen, Di and Yu, Feiwu and Zhao, Haiming and Yang, Jianxiao and others},
  journal={arXiv preprint arXiv:2503.20314},
  year={2025}
}

@article{transformer,
  title={Attention is all you need},
  author={Vaswani, Ashish and Shazeer, Noam and Parmar, Niki and Uszkoreit, Jakob and Jones, Llion and Gomez, Aidan N and Kaiser, {\L}ukasz and Polosukhin, Illia},
  journal={Advances in neural information processing systems},
  volume={30},
  year={2017}
}

@article{sana,
  title={Sana: Efficient high-resolution image synthesis with linear diffusion transformers},
  author={Xie, Enze and Chen, Junsong and Chen, Junyu and Cai, Han and Tang, Haotian and Lin, Yujun and Zhang, Zhekai and Li, Muyang and Zhu, Ligeng and Lu, Yao and others},
  journal={arXiv preprint arXiv:2410.10629},
  year={2024}
}

@article{npo,
  title={Diffusion-npo: Negative preference optimization for better preference aligned generation of diffusion models},
  author={Wang, Fu-Yun and Shui, Yunhao and Piao, Jingtan and Sun, Keqiang and Li, Hongsheng},
  journal={arXiv preprint arXiv:2505.11245},
  year={2025}
}

@article{repa,
  title={Representation alignment for generation: Training diffusion transformers is easier than you think},
  author={Yu, Sihyun and Kwak, Sangkyung and Jang, Huiwon and Jeong, Jongheon and Huang, Jonathan and Shin, Jinwoo and Xie, Saining},
  journal={arXiv preprint arXiv:2410.06940},
  year={2024}
}

@inproceedings{gan_t2v,
  title={Video generation from text},
  author={Li, Yitong and Min, Martin and Shen, Dinghan and Carlson, David and Carin, Lawrence},
  booktitle={Proceedings of the AAAI conference on artificial intelligence},
  volume={32},
  number={1},
  year={2018}
}

@inproceedings{TGans_c,
  title={To create what you tell: Generating videos from captions},
  author={Pan, Yingwei and Qiu, Zhaofan and Yao, Ting and Li, Houqiang and Mei, Tao},
  booktitle={Proceedings of the 25th ACM international conference on Multimedia},
  pages={1789--1798},
  year={2017}
}

@article{svd,
  title={Stable video diffusion: Scaling latent video diffusion models to large datasets},
  author={Blattmann, Andreas and Dockhorn, Tim and Kulal, Sumith and Mendelevitch, Daniel and Kilian, Maciej and Lorenz, Dominik and Levi, Yam and English, Zion and Voleti, Vikram and Letts, Adam and others},
  journal={arXiv preprint arXiv:2311.15127},
  year={2023}
}

@article{animatediff,
  title={Animatediff: Animate your personalized text-to-image diffusion models without specific tuning},
  author={Guo, Yuwei and Yang, Ceyuan and Rao, Anyi and Liang, Zhengyang and Wang, Yaohui and Qiao, Yu and Agrawala, Maneesh and Lin, Dahua and Dai, Bo},
  journal={arXiv preprint arXiv:2307.04725},
  year={2023}
}

@article{Magicvideo,
  title={Magicvideo: Efficient video generation with latent diffusion models},
  author={Zhou, Daquan and Wang, Weimin and Yan, Hanshu and Lv, Weiwei and Zhu, Yizhe and Feng, Jiashi},
  journal={arXiv preprint arXiv:2211.11018},
  year={2022}
}

@article{Lavie,
  title={Lavie: High-quality video generation with cascaded latent diffusion models},
  author={Wang, Yaohui and Chen, Xinyuan and Ma, Xin and Zhou, Shangchen and Huang, Ziqi and Wang, Yi and Yang, Ceyuan and He, Yinan and Yu, Jiashuo and Yang, Peiqing and others},
  journal={International Journal of Computer Vision},
  volume={133},
  number={5},
  pages={3059--3078},
  year={2025},
  publisher={Springer}
}

@inproceedings{scfg,
  title={Rethinking the spatial inconsistency in classifier-free diffusion guidance},
  author={Shen, Dazhong and Song, Guanglu and Xue, Zeyue and Wang, Fu-Yun and Liu, Yu},
  booktitle={Proceedings of the IEEE/CVF Conference on Computer Vision and Pattern Recognition},
  pages={9370--9379},
  year={2024}
}

@article{icg,
  title={No training, no problem: Rethinking classifier-free guidance for diffusion models},
  author={Sadat, Seyedmorteza and Kansy, Manuel and Hilliges, Otmar and Weber, Romann M},
  journal={arXiv preprint arXiv:2407.02687},
  year={2024}
}

@article{seg,
  title={Smoothed energy guidance: Guiding diffusion models with reduced energy curvature of attention},
  author={Hong, Susung},
  journal={Advances in Neural Information Processing Systems},
  volume={37},
  pages={66743--66772},
  year={2024}
}

@article{nag,
  title={Normalized Attention Guidance: Universal Negative Guidance for Diffusion Model},
  author={Chen, Dar-Yen and Bandyopadhyay, Hmrishav and Zou, Kai and Song, Yi-Zhe},
  journal={arXiv preprint arXiv:2505.21179},
  year={2025}
}

@inproceedings{vbench,
  title={Vbench: Comprehensive benchmark suite for video generative models},
  author={Huang, Ziqi and He, Yinan and Yu, Jiashuo and Zhang, Fan and Si, Chenyang and Jiang, Yuming and Zhang, Yuanhan and Wu, Tianxing and Jin, Qingyang and Chanpaisit, Nattapol and others},
  booktitle={Proceedings of the IEEE/CVF Conference on Computer Vision and Pattern Recognition},
  pages={21807--21818},
  year={2024}
}

@article{Vbench++,
  title={Vbench++: Comprehensive and versatile benchmark suite for video generative models},
  author={Huang, Ziqi and Zhang, Fan and Xu, Xiaojie and He, Yinan and Yu, Jiashuo and Dong, Ziyue and Ma, Qianli and Chanpaisit, Nattapol and Si, Chenyang and Jiang, Yuming and others},
  journal={IEEE Transactions on Pattern Analysis and Machine Intelligence},
  year={2025},
  publisher={IEEE}
}

@inproceedings{Evalcrafter,
  title={Evalcrafter: Benchmarking and evaluating large video generation models},
  author={Liu, Yaofang and Cun, Xiaodong and Liu, Xuebo and Wang, Xintao and Zhang, Yong and Chen, Haoxin and Liu, Yang and Zeng, Tieyong and Chan, Raymond and Shan, Ying},
  booktitle={Proceedings of the IEEE/CVF Conference on Computer Vision and Pattern Recognition},
  pages={22139--22149},
  year={2024}
}

@inproceedings{webvid,
  title={Frozen in time: A joint video and image encoder for end-to-end retrieval},
  author={Bain, Max and Nagrani, Arsha and Varol, G{\"u}l and Zisserman, Andrew},
  booktitle={Proceedings of the IEEE/CVF international conference on computer vision},
  pages={1728--1738},
  year={2021}
}

@article{Modelscope,
  title={Modelscope text-to-video technical report},
  author={Wang, Jiuniu and Yuan, Hangjie and Chen, Dayou and Zhang, Yingya and Wang, Xiang and Zhang, Shiwei},
  journal={arXiv preprint arXiv:2308.06571},
  year={2023}
}

@article{VideoCrafter,
  title={Latent video diffusion models for high-fidelity video generation with arbitrary lengths. arXiv 2022},
  author={He, Y and Yang, T and Zhang, Y and Shan, Y and Chen, Q},
  journal={arXiv preprint arXiv:2211.13221}
}

@article{Cogvideo,
  title={Cogvideo: Large-scale pretraining for text-to-video generation via transformers},
  author={Hong, Wenyi and Ding, Ming and Zheng, Wendi and Liu, Xinghan and Tang, Jie},
  journal={arXiv preprint arXiv:2205.15868},
  year={2022}
}
\bibliographystyle{icml2026}

%%%%%%%%%%%%%%%%%%%%%%%%%%%%%%%%%%%%%%%%%%%%%%%%%%%%%%%%%%%%%%%%%%%%%%%%%%%%%%%
%%%%%%%%%%%%%%%%%%%%%%%%%%%%%%%%%%%%%%%%%%%%%%%%%%%%%%%%%%%%%%%%%%%%%%%%%%%%%%%
% APPENDIX
%%%%%%%%%%%%%%%%%%%%%%%%%%%%%%%%%%%%%%%%%%%%%%%%%%%%%%%%%%%%%%%%%%%%%%%%%%%%%%%
%%%%%%%%%%%%%%%%%%%%%%%%%%%%%%%%%%%%%%%%%%%%%%%%%%%%%%%%%%%%%%%%%%%%%%%%%%%%%%%
\newpage
\appendix

\section{Implementation and Training Details}

\subsection{Implementation Details}
To ensure a fair and accurate evaluation, we strictly keep all training resources and configurations consistent across experiments, including computational budget and the total number of iterations. Specifically, all models are trained under the Fully Sharded Data Parallel (FSDP) framework using 16 H20 GPUs. During training, we adopt a linear warm-up schedule with a learning rate of 1.5e-4 and apply gradient clipping to prevent training instability. Each model is trained for 30,000 iterations with an identical amount of training data to ensure comparability.

\subsection{Training Details}
\label{sec:training_paradigm}
\begin{figure}[h]
\centering
\includegraphics[width=\linewidth]{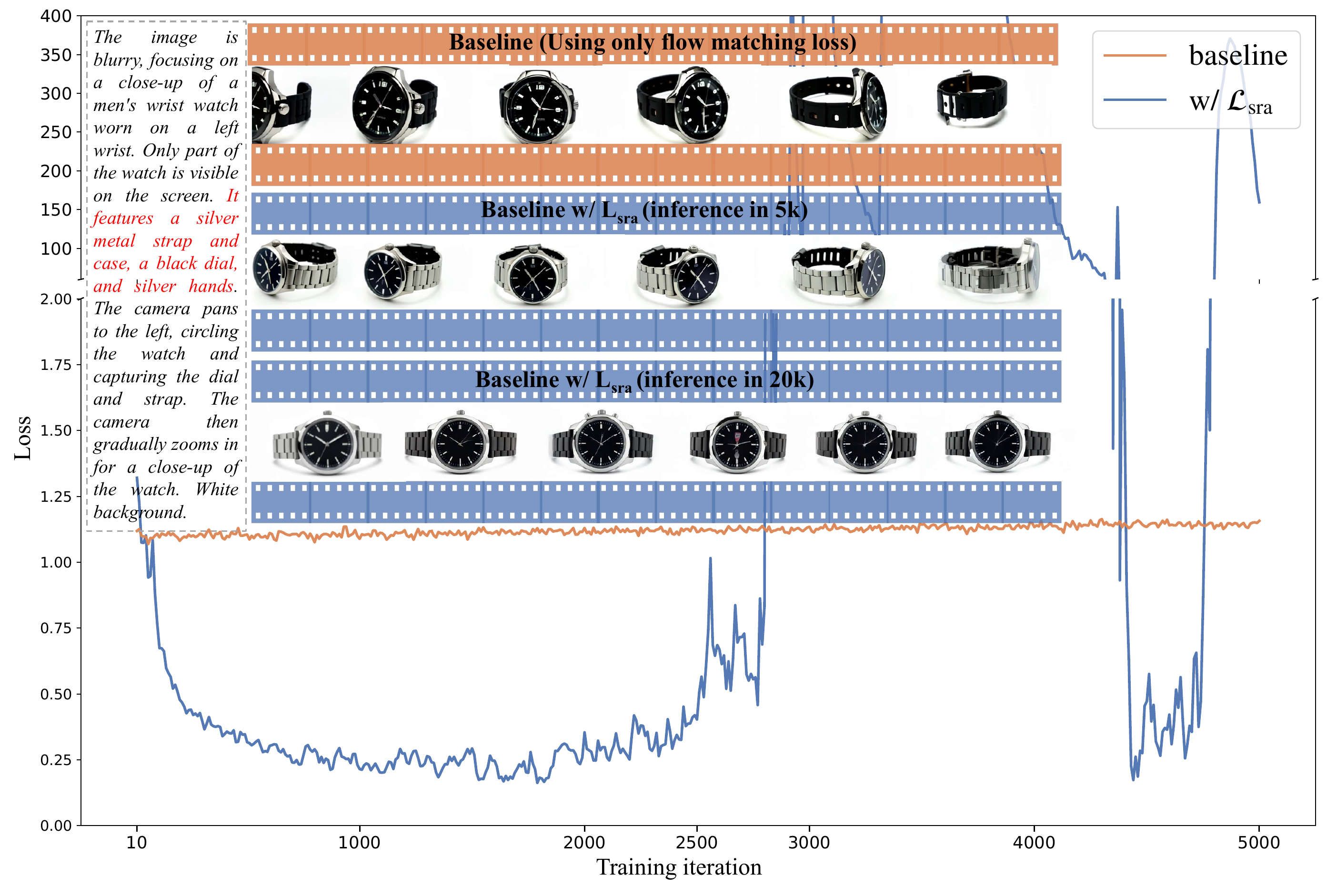} 
\caption{The loss curve during training is shown. Introducing $L_{sra}$ accelerates the model's fitting of the conditional semantic representation, enhancing the semantic consistency between the video and the conditional text. However, in the later stages of training, this leads to oscillations in the loss, causing model instability and a shift in focus towards the consistency between the conditional semantics and individual frame images.}
    \label{fig:train_loss}
\end{figure}

During model training, we initially assigned equal weights to $L_{dr}$ and $L_{sra}$. Observations indicated that in the early stages, $L_{sra}$ gradually converged, effectively enhancing the model's ability to fit prompts and accelerating the learning of specific token information. However, during the mid-to-late stages, $L_{sra}$ began to oscillate, leading to model collapse as iterations increased. As shown in Fig.~\ref{fig:train_loss}, introducing SRA for token-level spatial representation alignment effectively strengthens token-level guidance, enabling the model to achieve improved semantic adherence with a minimal number of training steps. However, as training progresses, the model becomes unstable and eventually collapses, leading to the loss of temporal consistency across video frames. Instead, the model tends to overfit the relationship between individual frames and the conditional text prompts. We hypothesize that using Mean Squared Error (MSE) for spatial representation constraints in $L_{sra}$ generates larger gradients. This causes the model to prioritize optimization gains from the Attention mechanism while neglecting the Flow Loss ($L_{dr}$). To address this, we developed an adaptive weighting strategy based on training steps. This strategy prioritizes the spatial constraints of $L_{sra}$ in the early stages to accelerate convergence, then transitions to $L_{dr}$ dominance in the later stages to correct the distribution fitting, encouraging the model to learn distribution predictions guided by complex semantic tokens.
\begin{equation}
\begin{cases}
      \alpha = max(0.0, \frac{1}{1+e^(0.001*(I - \hbar))}), \\
      L_{dare}= \alpha \cdot L_{sra} + (1-\alpha) \cdot L_{dr}
\end{cases}
\label{eq:final_lossh}
\end{equation}
In the above equation, $I$ denotes the number of training iterations, and $\hbar$ represents the position of the $L_{sra}$ oscillation point.

\subsection{Inference Details}
During inference, we use 8 NVIDIA H20 GPUs and strictly follow the same inference configuration as Wan 2.1. The sampling process runs for 40 steps with a classifier-free guidance (CFG) scale of 5.0. The negative prompt is set to:
"\textit{Worst quality, Normal quality, Low quality, Low res, Blurry, Jpeg artifacts, Grainy, text, logo, watermark, banner, extra digits, signature, subtitling, Bad anatomy, Bad proportions, Deformed, Disconnected limbs, Disfigured, Extra arms, Extra limbs, Extra hands, Fused fingers, Gross proportions, Long neck, Malformed limbs, Mutated, Mutated hands, Mutated limbs, Missing arms, Missing fingers, Poorly drawn hands, Poorly drawn face, Nsfw, Uncensored, Cleavage, Nude, Nipples, Overexposed, Plain background, Grainy, Underexposed, Deformed structures.}"

\section{Experiments}
\begin{table*}[h!]
\centering
\resizebox{\textwidth}{!}{
\begin{tabular}{ccccccc}
\hline
\multirow{2}{*}{Method} & \multicolumn{6}{c}{VBench} \\ \cline{2-7} 
 & \begin{tabular}[c]{@{}c@{}}Imaging\\ Quality(\%)\end{tabular} $\uparrow$ & \begin{tabular}[c]{@{}c@{}}Dynamic\\ Degree(\%)\end{tabular} $\uparrow$ & \begin{tabular}[c]{@{}c@{}}Aesthetic\\ Quality(\%)\end{tabular} $\uparrow$ & \begin{tabular}[c]{@{}c@{}}Subject\\ Consistency(\%)\end{tabular} $\uparrow$ & \begin{tabular}[c]{@{}c@{}}Background\\ Consistency(\%)\end{tabular} $\uparrow$ & \begin{tabular}[c]{@{}c@{}}Motion\\ Smoothness(\%)\end{tabular} $\uparrow$ \\ \hline
Wan 2.1    & 58.01    &  96.00   &  45.28  &  89.61   &  92.01 & 98.04 \\
\rowcolor{IceBlue!80} w. DARE (Ours)  & 63.12 & 84.00 & 44.50 & 95.19 & 95.64 & 99.19 \\  \hline
Seedance1.0\_exp   & 69.55    & 94.00   & 49.56      & 93.32     & 94.52    & 98.44     \\
\rowcolor{IceBlue!80} w. DARE (Ours)   &  69.29  &  89.00    &  49.79     &  93.39   &   94.65   &  98.46  \\ \hline
\end{tabular}
}
\resizebox{\textwidth}{!}{
\begin{tabular}{cccccccccc}
\hline
\multirow{2}{*}{Method} & \multicolumn{9}{c}{EvalCrafter}  \\ \cline{2-10} 
& \begin{tabular}[c]{@{}c@{}}CLIP\\ Temp\end{tabular} $\uparrow$ & \begin{tabular}[c]{@{}c@{}}SD\\ Score\end{tabular} $\uparrow$ & \begin{tabular}[c]{@{}c@{}}CLIP\\ Score\end{tabular} $\uparrow$ & \begin{tabular}[c]{@{}c@{}}Face\\ Consistency\end{tabular} $\uparrow$ & \begin{tabular}[c]{@{}c@{}}BLIP\\ BLUE\end{tabular} $\uparrow$ & IS   $\uparrow$ & IS$_{std}$ $\downarrow$    & \begin{tabular}[c]{@{}c@{}}Flow\\ Score\end{tabular} & \begin{tabular}[c]{@{}c@{}}Warping\\ Error\end{tabular} $\downarrow$ \\ \hline
Wan 2.1  & 0.9970  & 0.667 & 0.222 &   0.987 & 0.032 & 6.63 & 1.12 & 5.81 & 243.1 \\
\rowcolor{IceBlue!80} w. DARE (Ours) & 0.9992    &  0.675      &  0.231        & 0.993  &  0.035  & 6.08  & 0.75 & 3.18  & 102.9  \\ \hline
Seedance1.0\_exp & 0.9975  & 0.661 & 0.218   & 0.985   & 0.028 & 6.54 & 0.73 & 2.43   & 212.4  \\
\rowcolor{IceBlue!80} w. DARE (Ours) &  0.9975  & 0.665 & 0.219   &  0.985    & 0.028 & 6.61  & 0.83 & 2.35  & 218.7   \\ \hline
\end{tabular}
}
\caption{Evaluation results of generated outputs under the VBench \cite{vbench,Vbench++} and EvalCrafter \cite{Evalcrafter} frameworks using self-designed evaluation prompts.}
\label{tab:eval_in_self_prompt}
\end{table*}

\subsection{Datasets}
WebVid \cite{webvid} is a large-scale web video–text alignment dataset containing over 10 million video–caption pairs. Compared with traditional manually annotated datasets, it offers much larger scale and richer semantic diversity, covering a wide range of scenarios. This makes it well suited for pretraining and fine-tuning models for video generation and cross-modal alignment. Since the captions are collected from the web, the dataset inevitably contains noise, which can help improve model robustness and generalization ability during training. During model performance evaluation, we generate videos using two sets of textual prompts. The first set consists of 100 prompts constructed with the assistance of GPT, while the second set contains 946 prompts provided by VBench \cite{vbench,Vbench++}.

\subsection{Evaluation Metrics}
To comprehensively evaluate the video generation capability of our model, we conduct a thorough assessment of the generated results using the EvalCrafter \cite{Evalcrafter} and VBench \cite{vbench,Vbench++} evaluation frameworks.
EvalCrafter provides a fine-grained and multi-dimensional evaluation by considering several key aspects, including temporal consistency, text–video alignment, image quality and diversity, and inter-frame continuity. This framework enables detailed analysis of different components that contribute to video generation quality.
VBench offers a more comprehensive evaluation protocol. It incorporates additional factors such as video aesthetic quality, motion dynamics, subject consistency, semantic alignment between the video and the input description, and global video coherence, thereby further extending the evaluation scope and enabling a more holistic assessment of generation quality.
Furthermore, this benchmark provides dimension-specific video prompts tailored to different evaluation aspects, allowing the model’s generation performance to be assessed more precisely and in a more targeted manner from multiple perspectives.

\subsection{Performance evaluation on the self-constructed benchmark dataset}
To further validate the effectiveness of our method, we additionally used GPT to construct 100 instructions. These instructions are semantically more complex, closer to real-world scenarios, and better reflect the performance of the base models. Based on this instruction set, we conducted model inference and evaluated the generated results using two benchmarks, VBench \cite{vbench} and EvalCrafter \cite{Evalcrafter}.
As shown in Table \ref{tab:eval_in_self_prompt}, after introducing DARE, the performance improves across multiple evaluation aspects compared with Seedance \cite{seedance}. However, the improvement is relatively modest. In contrast, our method achieves significant performance gains on Wan \cite{wan}. This is mainly because existing evaluation metrics focus primarily on image quality and content consistency. Since Seedance already mitigates collapse issues to a certain extent, introducing DARE on top of it does not lead to substantial improvements under these metrics. For Wan, however, the base model may still suffer from various collapse or degradation issues. After integrating DARE, these issues are effectively suppressed, leading to improved spatiotemporal continuity and reduced collapse across frames.
These results indicate that our method, which performs distribution correction and spatial alignment from the perspective of tokens, not only enhances the model’s ability to learn semantic representations of different tokens but also effectively mitigates spatiotemporal collapse caused by abrupt inter-frame transitions.

\subsection{Performance evaluation on the VBench benchmark dataset}
Although the generated results have already been evaluated using the VBench benchmark, we further introduce an additional benchmark to provide a more comprehensive evaluation of our method. Specifically, we use EvalCrafter to assess the generation results produced with the same VBench prompts.
As shown in Table \ref{tab:eval_vbench_prompt_in_evalcrafter}, EvalCrafter primarily evaluates generated content from the perspectives of image quality and visual perception. We observe that, after introducing DARE, the model achieves consistent improvements in both visual perception alignment and semantic alignment. In addition, compared with the relatively weaker baseline Wan, incorporating our method not only yields advantages in semantic perception evaluation but also effectively prevents the occurrence of generation collapse.
\begin{table*}[ht!]
\centering
\resizebox{\textwidth}{!}{
\begin{tabular}{cccccccccc}
\hline
\multirow{2}{*}{Method} & \multicolumn{9}{c}{EvalCrafter}  \\ \cline{2-10} 
& \begin{tabular}[c]{@{}c@{}}CLIP\\ Temp\end{tabular} $\uparrow$ & \begin{tabular}[c]{@{}c@{}}SD\\ Score\end{tabular} $\uparrow$ & \begin{tabular}[c]{@{}c@{}}CLIP\\ Score\end{tabular} $\uparrow$ & \begin{tabular}[c]{@{}c@{}}Face\\ Consistency\end{tabular} $\uparrow$ & \begin{tabular}[c]{@{}c@{}}BLIP\\ BLUE\end{tabular} $\uparrow$ & IS   $\uparrow$ & IS$_{std}$ $\downarrow$    & \begin{tabular}[c]{@{}c@{}}Flow\\ Score\end{tabular} & \begin{tabular}[c]{@{}c@{}}Warping\\ Error\end{tabular} $\downarrow$ \\ \hline
Wan 2.1   & 0.9983 & 0.7135 & 0.2098 & 0.9923 & 0.2137 & 12.63 & 2.147 & 2.25 & 124.73 \\
\rowcolor{IceBlue!80}w. DARE (Ours) & 0.9988 & 0.7235 & 0.2150 & 0.9929  & 0.2154 & 11.11 & 3.193 & 1.56 & 71.32 \\ \hline
Seedance1.0\_exp      & 0.9986       & 0.7167    & 0.214         & 0.9909    & 0.2552    & 15.02 & 3.94 & 1.15     & 141.798  \\
\rowcolor{IceBlue!80}w. DARE (Ours) & 0.9986 & 0.7195 & 0.214  & 0.9907 & 0.2625 & 15.33  & 3.75  & 1.16 &   152.028  \\ \hline
\end{tabular}}
\caption{Evaluation results of generated outputs under the EvalCrafter \cite{Evalcrafter} frameworks using the prompts provided by VBench.}
\label{tab:eval_vbench_prompt_in_evalcrafter}
\end{table*}

\begin{figure*}[th!]
\centering
\begin{subfigure}[t]{0.73\linewidth}
    \centering
    \includegraphics[width=\linewidth]{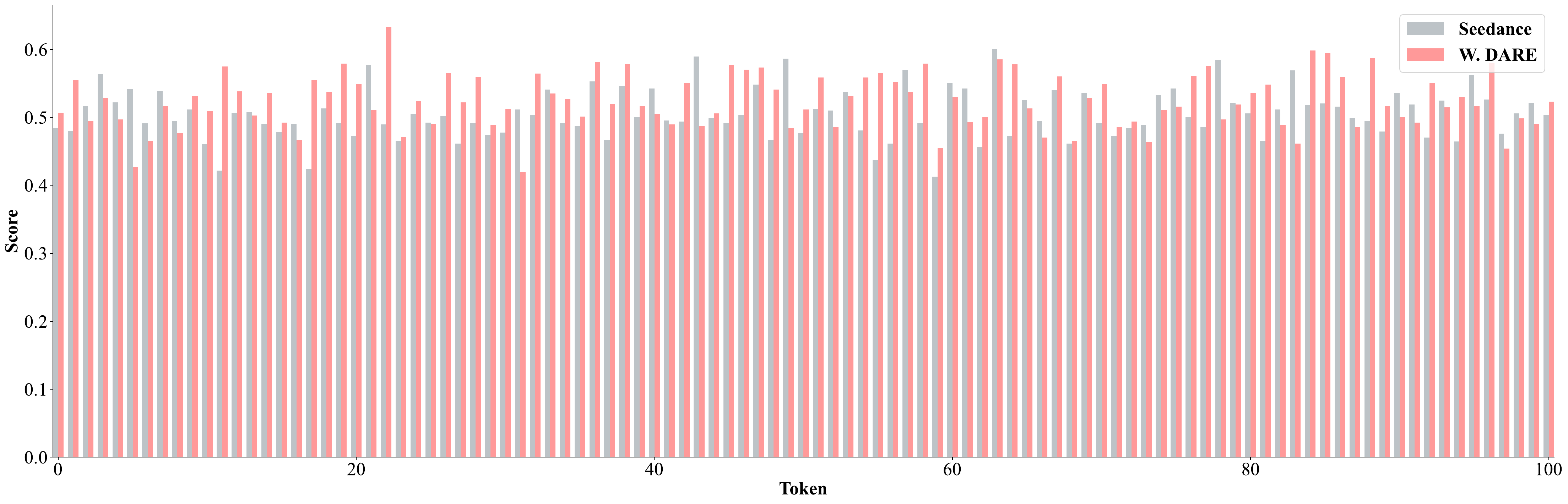}
    \caption{}
    \label{fig:token_score}
\end{subfigure}
\begin{subfigure}[t]{0.22\linewidth}
    \centering
    \includegraphics[width=\linewidth]{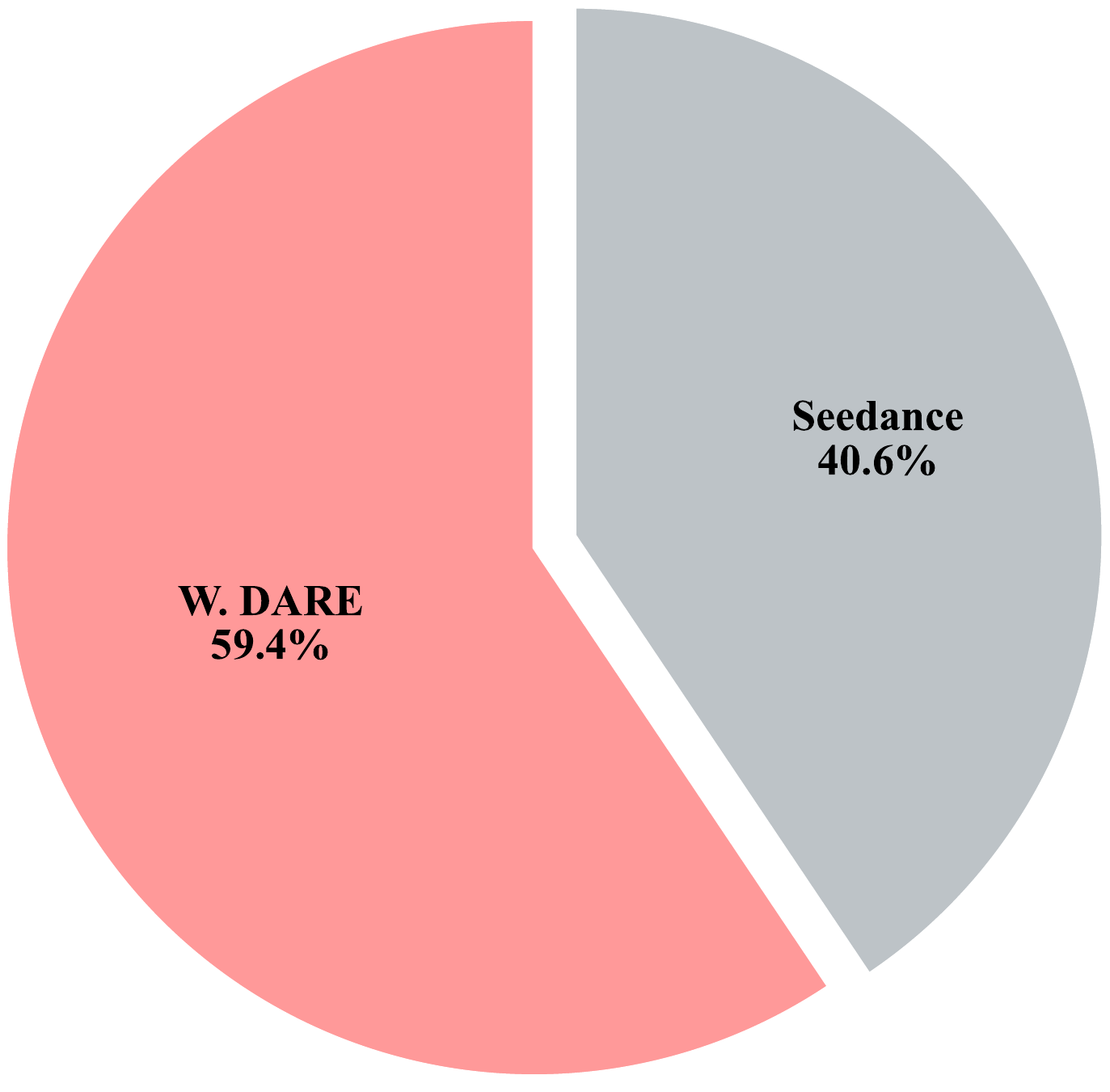}
    \caption{}
    \label{fig:perform_score}
\end{subfigure}
\begin{subfigure}[t]{\linewidth}
    \centering
    \includegraphics[width=0.97\linewidth]{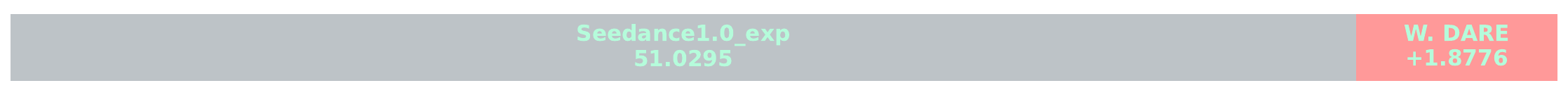}
    \caption{}
    \label{fig:overall_score}
\end{subfigure}
\caption{Visual comparison and analysis of the attention scores produced by high-density semantic tokens during the inference stage. Using 101 high-density semantic tokens extracted through statistical analysis, (a) illustrates the changes in the attention scores of different tokens before and after introducing DARE; (b) shows the proportion of tokens that exhibit a significant increase in attention scores after incorporating DARE; and (c) presents the overall variation in attention scores before and after applying DARE.}
\label{fig:token_score_detail}
\end{figure*}

\subsection{Attention Score–Based Visualization of High-Density Semantic Token Learning}
To further investigate the model's learning dynamics on tokens of varying semantic densities before and after integrating DARE, we sampled 101 high-semantic-density (i.e., high-importance-weight) tokens based on data stored during training. We then conducted single-token inference to extract the attention score for each token. As illustrated in Fig. ~\ref{fig:token_score_detail}, we analyze the results from three perspectives. First, Fig. ~\ref{fig:token_score} presents a visual comparison of the attention scores across all selected tokens. To explicitly highlight the overall shift in attention scores caused by DARE, Fig. ~\ref{fig:perform_score} illustrates the proportion of tokens exhibiting score improvements. Evidently, after incorporating DARE, approximately 59.4\% of the tokens demonstrated deepened learning, indicating enhanced attention allocation and distribution-fitting capabilities. Furthermore, Fig. ~\ref{fig:overall_score} compares the mean attention scores; the baseline average of 51.0295 increased by 1.8776 following the introduction of DARE.

In summary, the integration of DARE effectively bolsters the model's ability to learn high-semantic-density tokens and intensifies its focus on them, thereby mitigating the information loss typically caused by over-attending to low-density semantic tokens. This substantiates that DARE significantly elevates generation quality while ensuring the output aligns more faithfully with the conditional semantic information.

\subsection{More qualitative comparison and analysis}
To more clearly demonstrate the effect of DARE on generation quality, we conduct qualitative comparisons on two evaluation sets using two base models.

\begin{figure*}[ht!]
    \centering
    \includegraphics[width=\linewidth]{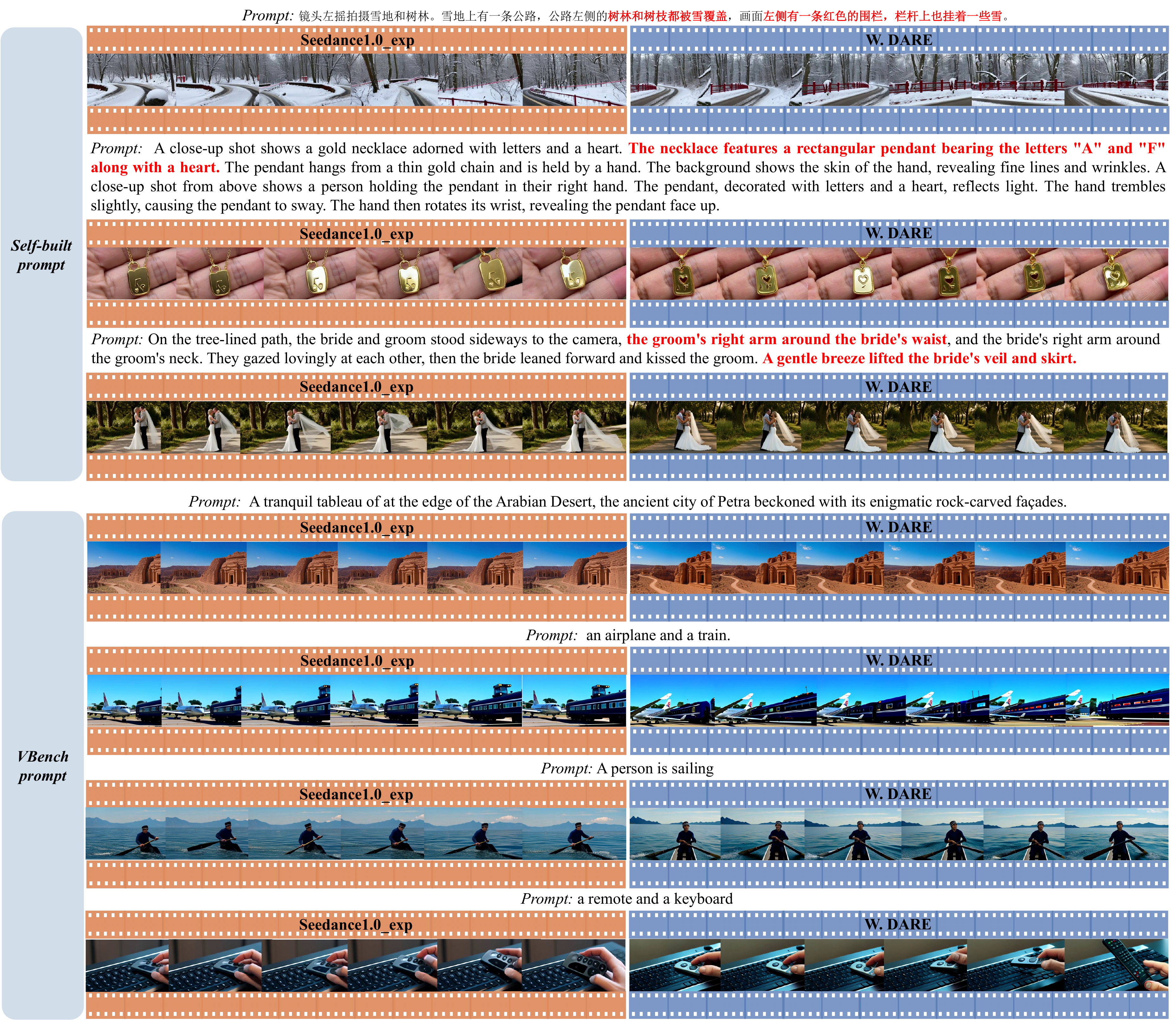}
    \caption{Qualitative comparison of generation results produced by Seedance before and after introducing DARE on the self-constructed dataset and the VBench benchmark.}
    \label{supplement:compare_seedance}
\end{figure*}
First, we present visual comparisons using Seedance \cite{seedance} on both a self-constructed dataset and the VBench benchmark. As shown in Fig. \ref{supplement:compare_seedance}, on the self-constructed dataset, Seedance tends to lose fine-grained semantic details. For example, in the second row, tokens such as “A” and “F” are ignored, while the model focuses primarily on the “rectangular pendant.” In contrast, by introducing token-level spatial consistency alignment, DARE preserves these fine-grained semantic details. Similarly, in the third row, Seedance fails to capture spatial positional relationships, leading to incorrect placement of the “groom” and “bride,” whereas DARE produces spatially consistent results.
On the VBench benchmark, although the prompts are relatively simple, the generated results often lack dynamic motion. For example, in the second row, the airplane and train remain largely static when generated by Seedance. After applying DARE, noticeable dynamic interaction between the objects emerges. In the third row, Seedance exhibits clear spatial misalignment, while DARE significantly improves spatial consistency and overall generation quality, producing results that better reflect realistic motion.

\begin{figure*}[ht!]
    \centering
    \includegraphics[width=\linewidth]{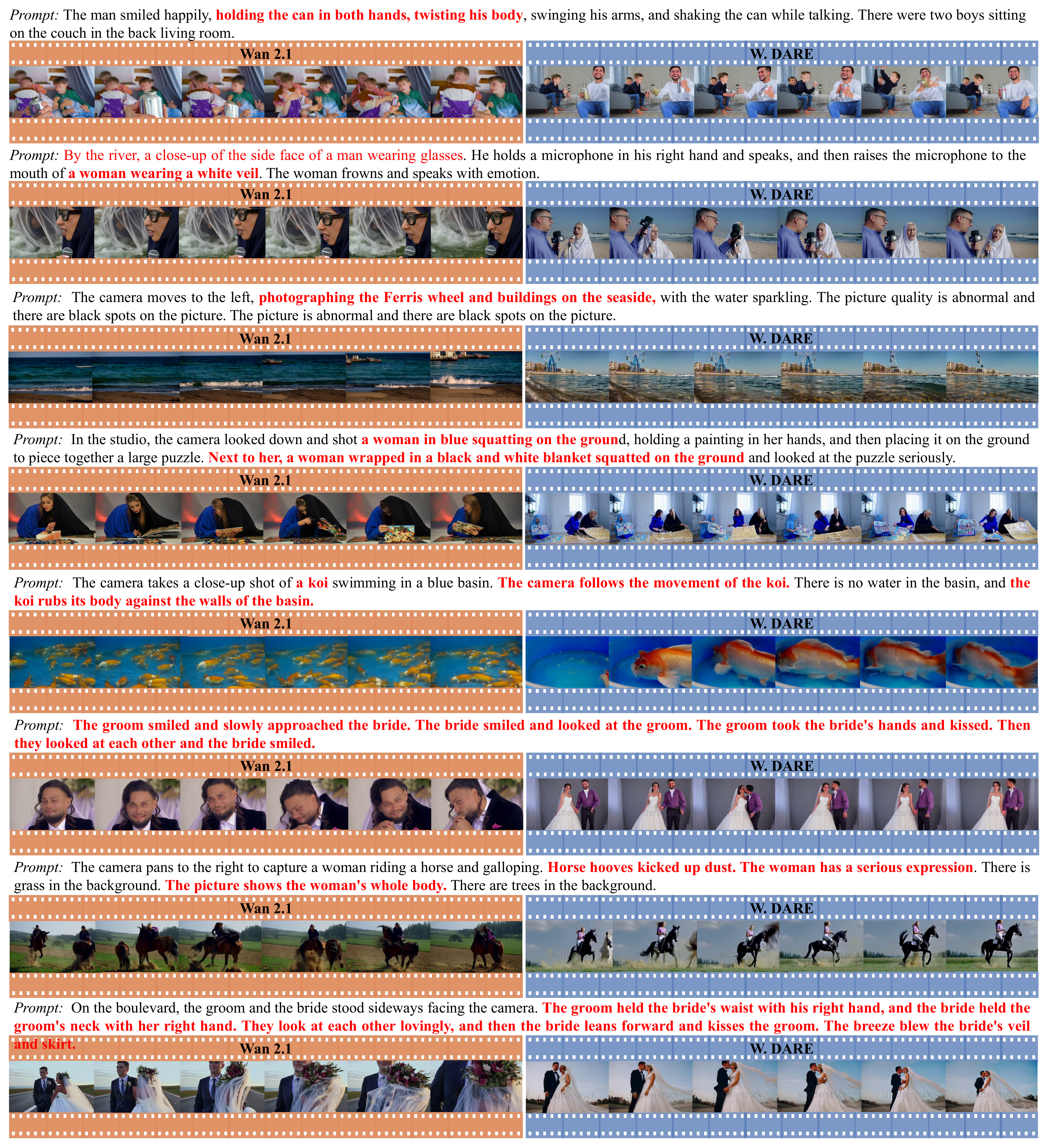}
    \caption{Qualitative comparison of generation results produced by Wan 2.1 before and after introducing DARE on the self-constructed dataset.}
    \label{supplement:compare_wan_ours_prompt}
\end{figure*}

Second, we perform qualitative comparisons using Wan 2.1 \cite{wan} on the self-constructed dataset. As shown in Fig. \ref{supplement:compare_wan_ours_prompt}, the base model often suffers from instability or collapse under complex semantic conditions. In the first row, the result generated by Wan 2.1 collapses and fails to capture spatial relationships, focusing only on a subset of tokens (e.g., “boy” and “can”) while ignoring other semantic cues. After introducing DARE, the collapse issue is largely mitigated and the generated content becomes highly consistent with the provided semantic guidance.
Similar collapse issues appear in rows 3, 4, 6, and 8, all of which are significantly improved with DARE. In rows 2, 5, and 7, the base model ignores certain semantic tokens, leading to clear inconsistencies between the generated content and the prompt. In contrast, DARE effectively improves semantic alignment while suppressing collapse.

\begin{figure*}[ht!]
    \centering
    \includegraphics[width=\linewidth]{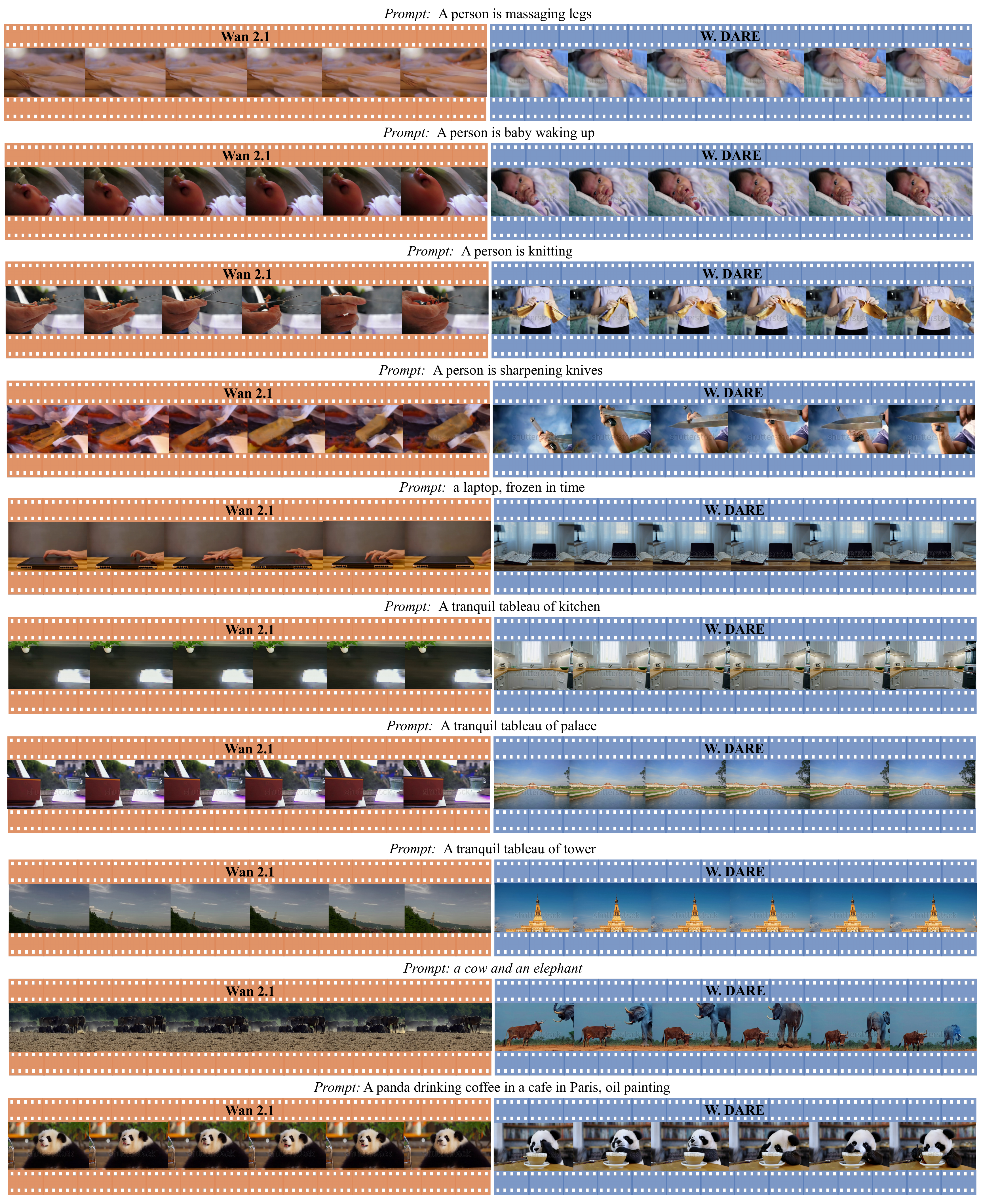}
    \caption{Qualitative comparison of generation results produced by Wan 2.1 before and after introducing DARE on the VBench benchmark.}
    \label{supplement:compare_wan_vbench_prompt}
\end{figure*}

Finally, we present qualitative comparisons using Wan 2.1 on the VBench benchmark (Fig. \ref{supplement:compare_wan_vbench_prompt}). Even with short and simple prompts, the base model may exhibit semantic misunderstanding or local collapse. For example, in rows 3 and 4, significant semantic misinterpretation leads to poor generation quality. In rows 1, 2, and 5, although the instructions are simple, distribution mismatch still results in localized collapse. With DARE, both generation quality and semantic consistency are significantly improved.

Overall, incorporating DARE into different base models consistently improves generation quality across multiple benchmarks. In particular, it enhances semantic fidelity while effectively mitigating generation collapse, demonstrating that the proposed components provide complementary solutions for improving conditional guidance and generation stability.

\end{document}